%% file: main.tex
\documentclass{article} 
\usepackage{iclr2025_conference,times}

\input{math_commands.tex}

\usepackage{hyperref}
\usepackage{url}

\usepackage{bm}
\usepackage{graphicx}
\usepackage{amsmath}
\usepackage{multirow}
\usepackage{booktabs}
\usepackage[normalem]{ulem}
\useunder{\uline}{\ul}{}

\usepackage{algorithm} 
\usepackage{algpseudocode}

\title{Channel-aware Contrastive Conditional Diffusion for Multivariate Probabilistic Time Series Forecasting}

\author{Siyang Li$^{1}$, \ Yize Chen$^{2}$, \ Hui Xiong$^{1}$\footnotemark[1]\thanks{Corresponding author.} \\ $^{1}$Hong Kong University of Science and Technology (Guangzhou)\\ $^{2}$University of Alberta\\
\texttt{sli572@connect.hkust-gz.edu.cn, yize.chen@ualberta.ca,} \\
\texttt{xionghui@hkust-gz.edu.cn} \\
}

\iclrfinalcopy 
\begin{document}

\maketitle
\begin{abstract}
\input{abstract}
\end{abstract}

\section{Introduction}
\input{introduction}

\section{Preliminaries}
\input{preliminaries}

\section{Method: Channel-aware contrastive conditional diffusion}
\input{methodology}

\section{Experiments}
\input{experiments}

\section{Related work}
\vspace{-5pt}
\input{related_work_short}

\section{Conclusion}
\vspace{-5pt}
\input{conclusion}

\subsubsection*{Ethics Statement}
Our work is only aimed at faithful multivariate probabilistic forecasting for human good, so there is no involvement of human subjects or conflict of interests as far as the authors are aware of.




\bibliography{iclr2025_conference}
\bibliographystyle{iclr2025_conference}

\newpage
\appendix
\section{Appendix}
\input{appendix}

\end{document}

%% file: math_commands.tex
\usepackage{amsmath,amsfonts,bm}

\def\eqref#1{equation~\ref{#1}}

\def\1{\bm{1}}

\DeclareMathAlphabet{\mathsfit}{\encodingdefault}{\sfdefault}{m}{sl}
\SetMathAlphabet{\mathsfit}{bold}{\encodingdefault}{\sfdefault}{bx}{n}



%% file: abstract.tex
Forecasting faithful trajectories of multivariate time series from practical scopes is essential for reasonable decision-making. Recent methods majorly tailor generative conditional diffusion models to estimate the target temporal predictive distribution. However, it remains an obstacle to enhance the exploitation efficiency of given implicit temporal predictive information to bolster conditional diffusion learning. To this end, we propose a generic channel-aware \underline{C}ontrastive \underline{C}onditional \underline{D}iffusion model entitled \textbf{CCDM} to achieve desirable \underline{M}ultivariate probabilistic forecasting, obviating the need for curated temporal conditioning inductive biases. In detail, we first design a channel-centric conditional denoising network to manage intra-variate variations and cross-variate correlations, which can lead to scalability on diverse prediction horizons and channel numbers. Then, we devise an ad-hoc denoising-based temporal contrastive learning to explicitly amplify the predictive mutual information between past observations and future forecasts. It can coherently complement naive step-wise denoising diffusion training and improve the forecasting accuracy and generality on unknown test time series. Besides, we offer theoretic insights on the benefits of such auxiliary contrastive training refinement from both neural mutual information and temporal distribution generalization aspects. The proposed CCDM can exhibit superior forecasting capability compared to current state-of-the-art diffusion forecasters over a comprehensive benchmark, with best MSE and CRPS outcomes on $66.67\%$ and $83.33\%$ cases. Our code is publicly available at 
\url{https://github.com/LSY-Cython/CCDM}.

%% file: introduction.tex
Multivariate probabilistic time series forecasting aims to quantify the stochastic temporal evolutions of multiple continuous variables and benefit decision-making in various engineering fields, such as weather prediction \citep{li2024generative}, renewable energy dispatch \citep{dumas2022deep}, traffic planning \citep{huang2023metaprobformer} and financial trading \citep{gao2024diffsformer}. Modern methods majorly customize time series generative models \citep{salinas2019high, li2022generative, yoon2019time, rasul2020multivariate} and produce diverse plausible trajectories to decipher the intricate temporal predictive distribution which is conditioned on past observations. Due to the excellent mode coverage capacity and training stability of diffusion models \citep{song2020score, ho2020denoising}, a flurry of conditional diffusion forecasters \citep{lin2023diffusion, yang2024survey} are recently developed by designing effective temporal conditioning mechanisms to discover informative patterns from historical time series.
 
Despite current advances, there remain two open challenges on learning a precise and generalizable multivariate predictive distribution via the step-wise denoising paradigm. The first obstacle is how to represent \textit{multivariate temporal correlations} in both past observed and target denoised sequences. To this end, many diffusion forecasting methods focus on \textit{efficient architectural designs} upon conditional denoising networks to capture heterogeneous temporal correlations in both diffusion corrupted and pure conditioning time series. Among them, spatiotemporal attention layers in CSDI \citep{tashiro2021csdi} and structured state space modules in SSSD \citep{alcaraz2022diffusion} are employed to characterize intra-channel and inter-channel \footnote{A channel shares the same physical interpretations with a variate, with each channel indicating a univariate time series.} relations. LDT \citep{feng2024latent} and D$^{3}$VAE \citep{li2022generative} utilize latent diffusion models to handle high-dimensional sequences. However, due to the limited capacity on identifying channel-specific and cross-channel properties, these temporal denoisers cannot manifest scalability and reliability on tough prediction tasks with numerous channels or long terms. Inspired by recent success of channel-centric views in multivariate point forecasting \citep{chen2024similarity, liu2023itransformer}, we propose to endow a \textit{unified channel manipulation strategy} into the conditional denoising network, as depicted in Fig. \ref{architecture}, where we stack channel-independent and channel-mixing modules to capture univariate variations and inter-variate correlations.

\begin{figure}[!t]
\vspace{-25pt}
\centering
\includegraphics[width=1.0\textwidth]{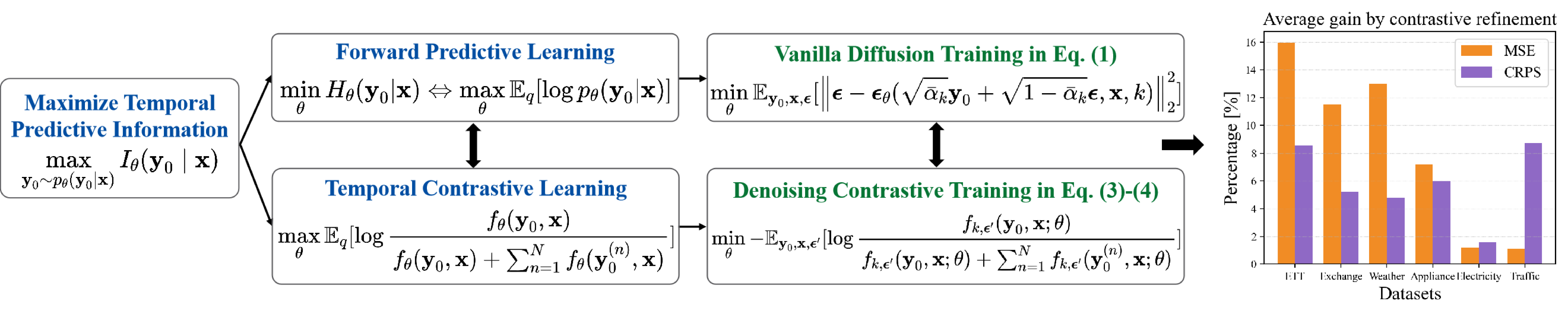}
\vspace{-15pt}
\caption{The schematic of proposed information-theoretic denoising-based contrastive diffusion learning. Bi-directional arrows indicate two learning ways are complementary. The bar chart depicts the average gains by contrastive diffusion refinement on diverse prediction setups for six datasets.}
\vspace{-10pt}
\label{motivation}
\end{figure}

The second issue is how to enhance the \textit{exploitation efficiency of the implicit predictive information} hidden in provided time series, which can improve the diversity and accuracy of generated profiles. It has been revealed that learning to unveil the useful temporal information like decomposed patterns \citep{deng2024bigger} or spectral biases \citep{crabbetime} can boost the estimated predictive distribution. Related diffusion forecasters also develop \textit{auxiliary training strategies} to amplify the fine-grained features, as the naive step-wise noise regression paradigm fall short in fully releasing the intrinsic predictive information. In particular, they employ specific temporal inductive biases to promote temporal conditioning schemes or guide iterative inference procedures. Pretraining temporal conditioning encoders by deterministic point prediction \citep{shen2023non, li2023transformer} is a viable method, 
which produces more accurate medians and sharper prediction intervals. Coupling unique temporal features like multi-granularity dynamics \citep{fan2024mg, shen2023multi} or target quantitative metrics \citep{kollovieh2024predict} to regularize the sequential diffusion process can also steer the reverse generation process towards plausible trajectories.
However, these auxiliary methods need to fully expose task-specific temporal properties or tailor distinct regulations for diffusion training and sampling procedures. They are neither consistent with standard hierarchical diffusion optimization nor generic over generative time series diffusion models.

Motivated by a neural information view in \citep{tsai2020self}, naive conditional time series diffusion learning can be deemed as a \textit{forward predictive way} to maximize the temporal mutual information between past observations and target forecasts. Such auxiliary inductive biases can empirically enrich the predictive temporal information. However, single predictive learning is inadequate to reveal the entire task-specific information. In light of the composite objective which integrates contrastive learning to procure more robust task-related representations \citep{tsai2020self}, we propose to further enhance the \textit{prediction-related mutual information} captured by denoising diffusion in a \textit{complementary contrastive way}, where both positive and negative time series are inspected at each diffusion step. We illustrate such temporal contrastive refinement on conditional diffusion forecasting in Fig. \ref{motivation}, which mitigates over-fitting and attains better generality on unknown test data.

In this work, we propose a contrastive conditional diffusion model termed CCDM which can explicitly maximize the predictive mutual information for multivariate probabilistic forecasting. The efficient \textit{channel-aware denoiser architecture} and complementary \textit{denoising-based contrastive refinement} are two recipes to boost diffusion forecasting capacity. Our main contributions are summarized as: (1) We design a composite channel-aware conditional denoising network, which merges channel-independent dense encoders to extract univariate dynamics and channel-wise diffusion transformers to aggregate cross-variate correlations. It gives rise to efficient iterative inference and better scalability on various channel numbers and prediction horizons. (2) We propose to explicitly amplify the predictive information between generated forecasts and past observations via a coherent denoising-based temporal contrastive learning, which can be seamlessly aligned with vanilla step-wise denoising diffusion training and thus efficient to implement. (3) Extensive simulations validate the superior forecasting capability of CCDM. It can attain better accuracy and reliability versus other excellent models on various forecasting settings, especially for long-term and large-channel scenarios.

%% file: preliminaries.tex
\subsection{Problem formulation}
In this paper, we look into the task of multivariate probabilistic time series forecasting. Given the past observation $\mathbf{x}\in \mathbb{R}^{L\times D}$ as conditioning time series, the goal is to generate a group of $S$ plausible forecasts $\{\hat{\mathbf{y}}_{0}^{(s)}\in \mathbb{R}^{H\times D}\}_{s=1}^{S}$ from the learned conditional predictive distribution $p_{\theta}(\mathbf{y}_{0}|\mathbf{x})$. Here, $D$ is the number of channels, $L$ and $H$ indicate the lookback window length and prediction horizon respectively. $\theta$ stands for the parameters of a conditional diffusion forecaster which represents the real predictive distribution $q(\mathbf{y}_{0}|\mathbf{x})$. We allocate diverse values to horizon $H$ and channel number $D$ to construct a holistic benchmark which can completely evaluate the capability of different conditional diffusion models on various forecasting scenarios.

\subsection{Conditional denoising diffusion models}
\label{subsec:cdm}
Conditional diffusion models have exhibited impressive capability on a wide variety of controllable multi-modal synthesis tasks \citep{chen2024overview}. It dictates a bi-directional distribution transport process between raw data $\mathbf{y}_{0}$ and prior Gaussian noise $\mathbf{y}_{K}\in \mathcal{N}(\mathbf{0}, \mathbf{I})$ via $K$ diffusion steps. The forward process gradually degrades clean $\mathbf{y}_{0}$ to fully noisy $\mathbf{y}_{K}$ and can be fixed as a Markov chain: $q(\mathbf{y}_{0:K})=q(\mathbf{y}_{0})\textstyle \prod_{k=1}^{K}q(\mathbf{y}_{k}|\mathbf{y}_{k-1})$, where $q(\mathbf{y}_{k}|\mathbf{y}_{k-1}):=\mathcal{N}(\mathbf{y}_{k}; \sqrt{1-\beta_{k}}\mathbf{y}_{k-1}, \beta_{k}\mathbf{I})$ and $\beta_{k}$ is the degree of imposed step-wise Gaussian noise. We can accelerate the forward sampling procedure and obtain closed-form latent state $\mathbf{y}_{k}$ at arbitrary step $k$ by a noteworthy property \citep{ho2020denoising}: $\mathbf{y}_{k}=\sqrt{\bar{\alpha}_{k}}\mathbf{y}_{0}+\sqrt{1-\bar{\alpha}_{k}}\bm{\epsilon}$, where $\bar{\alpha}_{k}:=\textstyle \prod_{s=1}^{k}(1-\beta_{s})$ and $\bm{\epsilon}\sim \mathcal{N}(\mathbf{0}, \mathbf{I})$. The reverse generation process converts known Gaussian to realistic prediction data $\mathbf{y}_{0}$ given input conditions $\mathbf{x}$, which can be cast as a parameterized Markov chain: $p_{\theta}(\mathbf{y}_{0:K}|\mathbf{x})=p(\mathbf{y}_{K})\textstyle \prod_{k=K}^{1}p_{\theta}(\mathbf{y}_{k-1}|\mathbf{y}_{k}, \mathbf{x})$. 
The overall training objective can be simplified as minimizing the step-wise denoising loss below:
\begin{equation}
\centering
\mathcal{L}_{k}^{denoise}=\mathbb{E}_{\mathbf{y}_{0},\mathbf{x},\bm{\epsilon}}[\left\| \bm{\epsilon}-\bm{\epsilon}_{\theta}(\sqrt{\bar{\alpha}_{k}}\mathbf{y}_{0}+\sqrt{1-\bar{\alpha}_{k}} \bm{\epsilon},\mathbf{x},k)\right\|_2^{2}].
\label{eq:1}
\end{equation}

A potential issue for current conditional diffusion models lies in forging an effective conditioning mechanism that can enhance the alignment between given conditions $\mathbf{x}$ and produced data $\mathbf{y}_{0}$, like the coherent semantics between textual descriptions and visual renderings \citep{esser2024scaling}, or the conformity of generated vehicle motions to scenario constraints \citep{jiang2023motiondiffuser}. However, such data consistency is hard to represent for temporal conditional probability modeling. We thus explicitly learn to amplify the prediction-related temporal information conveyed from past conditioning time series to generated trajectories. Such predictive mutual information can reflect underlying temporal properties in historical sequences, to which the produced forecasts should comply.

\subsection{Neural mutual information maximization}
\label{subsec:nmi}
As discussed above, to more efficiently represent the useful predictive modes involved in conditioning time series, we choose to explicitly maximize the prediction-oriented mutual information when learning the conditional diffusion forecaster. Learning to maximize mutual information is effective to boost the consistency between two associated variables \citep{song2019understanding}, which has been actively applied to self-supervised learning \citep{liang2024factorized} and multi-modal alignment \citep{liang2024quantifying}. Regarding conditional diffusion learning, there also exist several related works \citep{wang2023infodiffusion, zhu2022discrete} which explicitly employ mutual information maximization to enhance high-level semantic coherence between input prompts and generated samples.
While we propose a complementary way to equip the conditional diffusion forecaster with this tool to bolster the utilization of informative temporal patterns. Besides, we provide a distinct composite loss design and more concrete interpretations on the benefits of the contrastive scheme to ordinary conditional diffusion. 

Among the two practical methods to maximize the intractable mutual information \citep{tsai2020self}, contrastive learning aids to strengthen the association by discriminating intra-class from inter-class samples. Contrastive predictive coding \citep{oord2018representation} realizes such objective by optimizing the contrastive lower bound with low variance via the prevalent InfoNCE loss:
\begin{equation}
\centering
\mathcal{L}_{InfoNCE}=-\mathbb{E}_{(\mathbf{y}_{0},\mathbf{x})\sim q(\mathbf{y}_{0},\mathbf{x}), \mathbf{y}_{0}^{(n)}\sim q^{n}(\mathbf{y}_{0})}  [\log \frac{f(\mathbf{y}_{0},\mathbf{x})}{f(\mathbf{y}_{0},\mathbf{x})+ {\textstyle \sum_{n=1}^{N}f(\mathbf{y}_{0}^{(n)},\mathbf{x})} } ].
\label{eq:2}
\end{equation}
During each iteration, we create a set of $N$ negative samples via the negative construction operation $q^{n}(\mathbf{y}_{0})$ on positive data $\mathbf{y}_{0}$. $f(\mathbf{y}_{0},\mathbf{x})$ accounts for the density ratio $\frac{q(\mathbf{y}_{0}|\mathbf{x})}{q(\mathbf{y}_{0})}$ and can be \textit{any types of positive real functions}. This flexible form of the density ratio function offers a natural initiative of the following denoising-based contrastive conditional diffusion. 

\textit{Forward predictive learning} is another way to boost the inter-dependency by fully reconstructing target $\mathbf{y}_{0}$ conditioned on given $\mathbf{x}$. This reconstruction learning can be realized by learning a deterministic mapping or conditional generative model from $\mathbf{x}$ to $\mathbf{y}_{0}$. As $I(\mathbf{y}_{0};\mathbf{x})=H(\mathbf{y}_{0})-H(\mathbf{y}_{0}|\mathbf{x})$, and $H(\mathbf{y}_{0})$ is irrelevant to discovering the entanglement between $\mathbf{x}$ and $\mathbf{y}_{0}$, thereby maximizing $I(\mathbf{x};\mathbf{y}_{0})$ boils down to optimizing the predictive lower bound 
$-H(\mathbf{y}_{0}|\mathbf{x})=\mathbb{E}_{q(\mathbf{x},\mathbf{y}_{0})}[\log{p_{\theta}(\mathbf{y}_{0}|\mathbf{x})}]$, which is aligned with the likelihood-based objective of naive conditional diffusion learning. 
\citep{tsai2020self} claims that combining both predictive and contrastive learning tactics can significantly raise the quality of obtained task-related features. Accordingly, we equip vanilla conditional time series diffusion with a denoising-based InfoNCE contrastive loss to further boost the temporal predictive information between past conditions and future forecasts. A concise motivation of this information-theoretic contrastive diffusion forecasting is depicted in Fig. \ref{motivation}.

%% file: methodology.tex
\begin{figure}[!t]
\vspace{-20pt}
\centering
\includegraphics[width=1.0\textwidth]{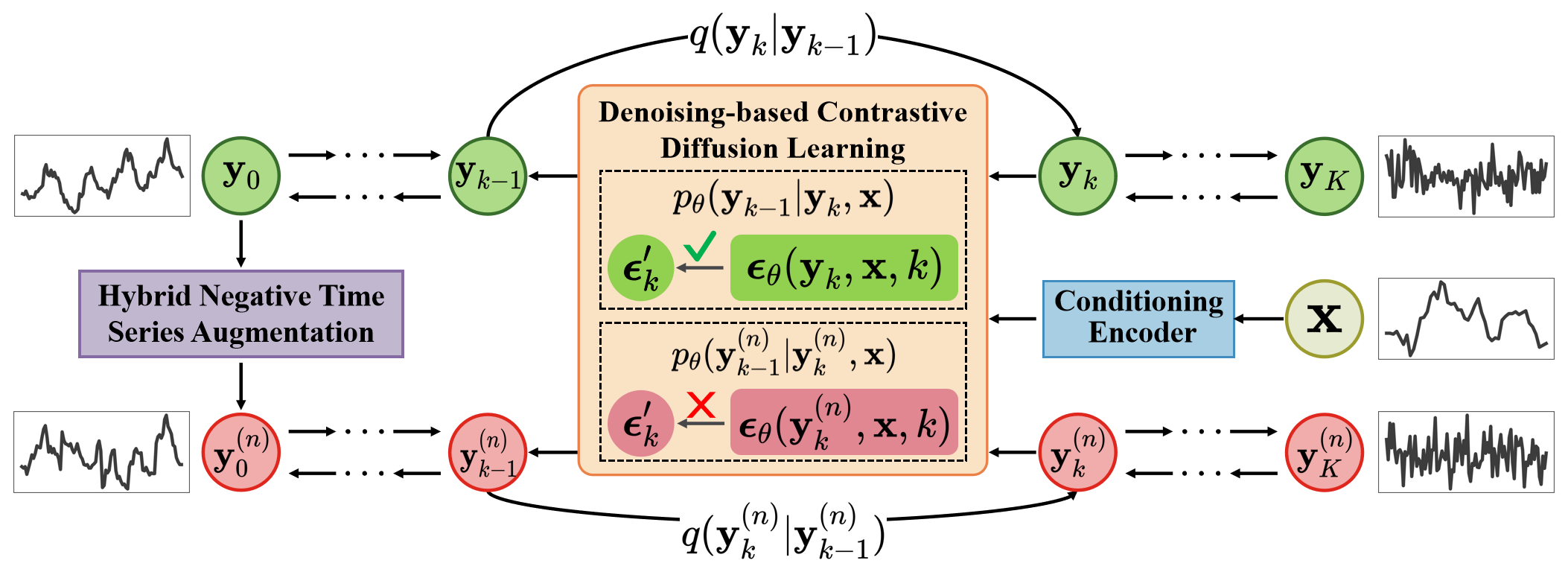}
\vspace{-10pt}
\caption{The framework of denoising-based contrastive conditional diffusion forecaster.}
\vspace{-10pt}
\label{framework}
\end{figure}

\begin{figure}[!t]
\vspace{-15pt}
\centering
\includegraphics[width=1.0\textwidth]{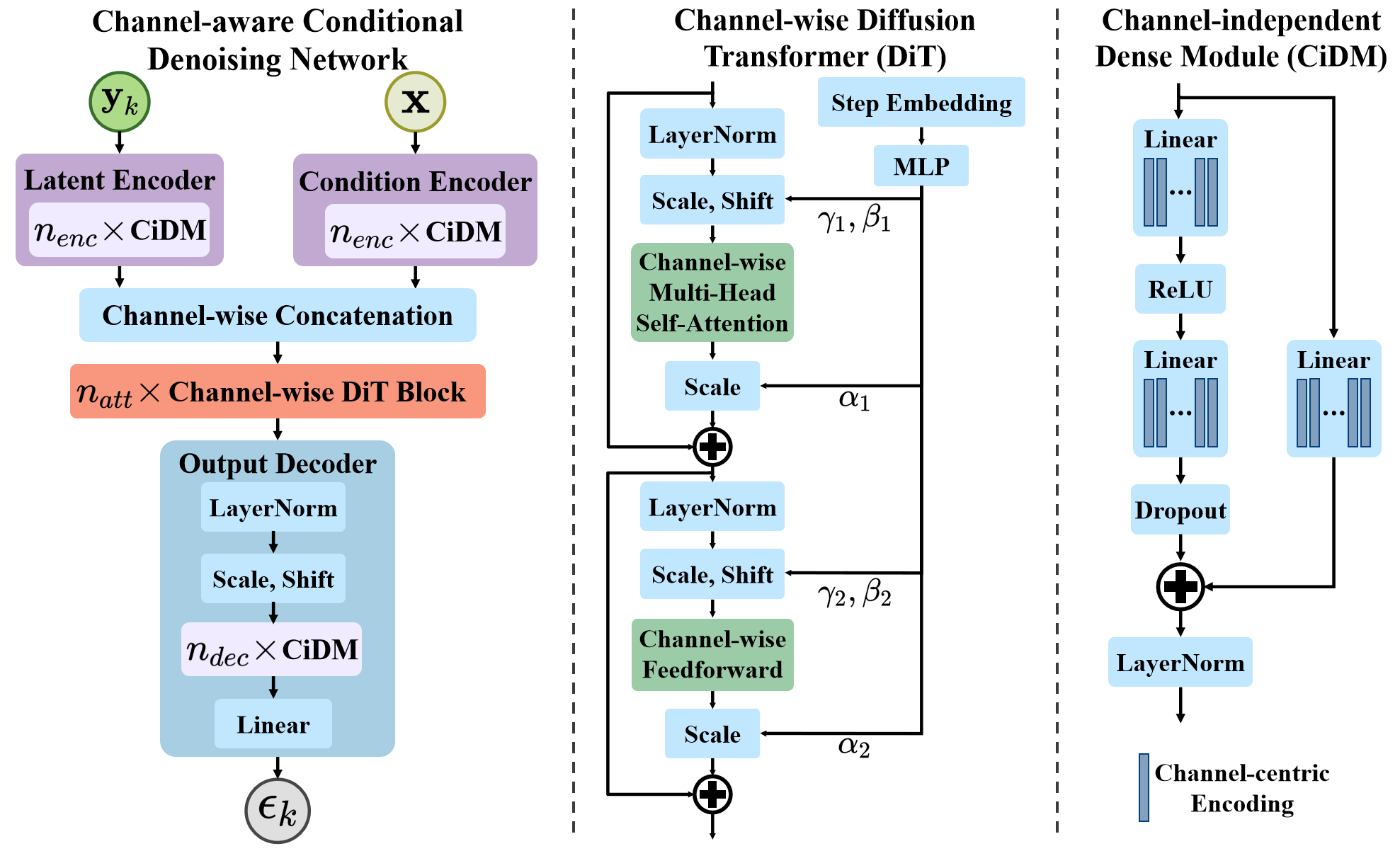}
\vspace{-10pt}
\caption{The diagram of channel-aware conditional denoiser architecture. \underline{Left}: the whole network. \underline{Middle}: channel-mixing DiT blocks. \underline{Right}: channel-independent MLP dense modules.}
\vspace{-10pt}
\label{architecture}
\end{figure}

In this section, we elucidate two innovations of the tailored CCDM for generative multivariate time series forecasting, including the hybrid channel-aware denoiser architecture depicted in Fig. \ref{architecture} and denoising-based contrastive diffusion learning demonstrated in Fig. \ref{framework}.

\subsection{Channel-aware conditional denoising network}
\label{subsec:denoiser}
Recent progress on multivariate prediction methods \citep{liu2023itransformer, ilbert2024unlocking} show that proper integration of channel management strategies in time series backbones is critical to discover univariate dynamics and cross-variate correlations. But this problem has not been well explored in multivariate diffusion forecasting and previous conditional denoiser structures do not obviously distinguish such heterogeneous channel-centric temporal properties. To this end, we design a channel-aware conditional denoising network which incorporates composite channel manipulation modules, i.e. channel-independent dense encoders and channel-mixing diffusion transformers. This architecture can efficiently represent intra-variate and inter-variate temporal correlations in past conditioning $\mathbf{x}$ and future predicted $\mathbf{y}_{0}$ under different noise levels, as well as being robust to diverse prediction horizons and channel numbers.

\textbf{Channel-independent dense encoders.}
We develop two channel-independent MLP encoders to extract unique temporal variations in each individual channel of observed condition $\mathbf{x}$ and corrupted latent state $\mathbf{y}_{k}$ at each diffusion step. The core ingredient in latent and condition encoders is the channel-independent dense module (CiDM) borrowed from TiDE \citep{das2023long}, which stands out as a potent MLP building-block for universal time series analysis models \citep{das2023decoder}. A salient element in CiDM is the skip-connecting MLP residual block which can improve temporal pattern expressivity. The $D$ linear layers in parallel are shared and used for separate channel feature embedding. We stack $n_{enc}$ CiDM modules of $e_{hid}$ hidden dimension to transform both $\mathbf{x}$ and $\mathbf{y}_{k}$ into $e_{hid}\times D$ size. These two input encoders can be easily adjusted to accommodate different context windows and hidden feature dimensions.

\textbf{Channel-wise diffusion transformers.} To regress step-wise Gaussian noise $\bm{\epsilon}_{k}$ on raw $\mathbf{y}_{0}$ more precisely, we should fully exploit implicit temporal information in pure conditioning $\mathbf{x}$ and polluted target $\mathbf{y}_{k}$. We concatenate the latent encoding of $\mathbf{x}$ and $\mathbf{y}_{k}$ along the channel axis and then leverage $n_{att}$-depth channel-wise diffusion transformer (DiT) blocks to aggregate heterogeneous temporal modes from various channels. DiT is an emergent diffusion backbone for open-ended text-to-image synthesis which merits eminent efficiency, scalability and robustness \citep{peebles2023scalable, esser2024scaling}. Two critical components in DiT are multi-head self-attention for feature fusion and adaptive layer norm (adaLN) layers to absorb other conditioning items (e.g. diffusion step embedding, text labels) as learnable scale and shift parameters. Although DiT has been adopted by few works \citep{cao2024timedit, feng2024latent} for generative time series modeling, they do not adapt it in a fully channel-centric angle. We repurpose DiT to model multivariate predictive distribution by replacing naive point-wise attention to a channel-wise version, which can blend correlated temporal information from different variates in $\mathbf{x}$ and $\mathbf{y}_{k}$.
Afterwards, we develop a output decoder with $n_{dec}$ CiDMs plus a last adaLN to yield the prediction of imposed noise $\bm{\epsilon}_{k}$ given $\mathbf{x}$ and $\mathbf{y}_{k}$.

\subsection{Denoising-based temporal contrastive refinement}
\label{subsec:contrastive}
Unlike previous empirically designed temporal conditioning schemes to make better exploitation of past predictive information, we instead propose to explicitly maximize the prediction-related mutual information $I(\mathbf{y}_{0}; \mathbf{x})$ between past observations $\mathbf{x}$ and future forecasts $\mathbf{y}_{0}$ via an adapted denoising-based contrastive strategy. We will employ the learnable denoising network $\bm{\epsilon}_{\theta}(\cdot)$ to represent the contrastive lower bound of $I(\mathbf{y}_{0}; \mathbf{x})$ presented by Eq. \ref{eq:2}, and exhibit this information-theoretic contrastive refinement is complementary and aligned with original conditional denoising diffusion optimization, which is actually another forward predictive method to maximize $I(\mathbf{y}_{0}; \mathbf{x})$.

To improve the diffusion forecasting capacity more essentially, the developed contrastive learning item is wished to directly benefit naive step-wise denoising-based training procedure, i.e. regularizing noise elimination behaviors of the conditional denoiser $\bm{\epsilon}_{\theta}(\cdot)$. Since the density ratio function $f(\mathbf{y}_{0},\mathbf{x})$ constituting the contrastive mutual information lower bound in Eq. \ref{eq:2} can be any positive-valued forms, this flexibility naturally motivates us to prescribe $f(\cdot)$ using the step-wise denoising objective in Eq. \ref{eq:1}, for both positive sample $\mathbf{y}_{0}$ and a group of negative samples $\mathbf{y}_{0}^{(n)}$:
\begin{subequations}
\begin{align}
\centering
f_{k,\bm{\epsilon}{}'}(\mathbf{y}_{0},\mathbf{x};\theta)&=\exp (-||\bm{\epsilon}{}'-\bm{\epsilon}_{\theta}(\sqrt{\bar{\alpha}_{k}}\mathbf{y}_{0}+\sqrt{1-\bar{\alpha}_{k}} \bm{\epsilon}{}' ,\mathbf{x},k)||^{2}_{2}/\tau);
\label{eq:3a}\\
f_{k,\bm{\epsilon}{}'}(\mathbf{y}_{0}^{(n)},\mathbf{x};\theta)&=\exp (-||\bm{\epsilon}{}'-\bm{\epsilon}_{\theta}(\sqrt{\bar{\alpha}_{k}}\mathbf{y}_{0}^{(n)}+\sqrt{1-\bar{\alpha}_{k}} \bm{\epsilon}{}' ,\mathbf{x},k)||^{2}_{2}/\tau);
\label{eq:3b}
\end{align}
\end{subequations}
where $\tau$ is the temperature coefficient in the softmax-form contrastive loss. The negative sample set $\{ \mathbf{y}_{0}^{(n)}\}_{n=1}^{N}$ is constructed by a hybrid time series augmentation method which alters both temporal variations and point magnitudes (See Appendix \ref{neg_aug} for details.). Then, we can derive the following contrastive refinement loss which is coincident with vanilla step-wise denoising diffusion training:
\begin{equation}
\centering
\mathcal{L}_{k}^{contrast}=-\mathbb{E}_{\mathbf{x},\mathbf{y}_{0},\{\mathbf{y}_{0}^{(n)}\}_{n=1}^{N},\bm{\epsilon}{}'}[\log \;\frac{f_{k,\bm{\epsilon}{}'}(\mathbf{y}_{0},\mathbf{x};\theta)}{f_{k,\bm{\epsilon}{}'}(\mathbf{y}_{0},\mathbf{x};\theta)+ {\textstyle \sum_{n=1}^{N}f_{k,\bm{\epsilon}{}'}(\mathbf{y}_{0}^{(n)},\mathbf{x};\theta)} } ].
\label{eq:4}
\vspace{-5pt}
\end{equation}

Apparently, the devised denoising-based temporal contrastive learning can not only seamlessly coordinate with standard diffusion training at each step $k$, but also improve the conditional denoiser behaviors in out-of-distribution (OOD) regions. These OOD areas are constituted by the low-density diffusion paths of negative samples, which are not touched by merely executing denoising learning along the high-density probability paths of positive samples.

\subsection{Overall learning objective}
The naive denoising diffusion model trained by log-likelihood maximization \citep{ho2020denoising} totally owns $K$-step valid training items. To align with this step-wise denoising distribution learning, we can amortize the contrastive regularization in Eq. \ref{eq:4} to each training step, and derive the overall learning objective below:
\begin{equation}
\centering
\max_{\theta} \;\mathbb{E}_{q(\mathbf{y}_{0}, \mathbf{x})}\left [ \log p_{\theta}(\mathbf{y}_{0}|\mathbf{x})+\lambda K \cdot I_{\theta}(\mathbf{y}_{0}; \mathbf{x}) \right ],
\label{eq:5}
\vspace{-5pt}
\end{equation}
where $\log p_{\theta}(\mathbf{y}_{0}|\mathbf{x})$ can be decomposed as $ \textstyle \sum_{k=1}^{K} \mathcal{L}_{k}^{denoise}$ and indicates the predictive distribution learning. Whilst $\max_{\theta} I_{\theta}(\mathbf{y}_{0}; \mathbf{x})$ governs the information-theoretic contrastive learning. Then, the practical training loss of the devised CCDM at each diffusion step can be presented as:
\begin{equation}
\centering
\mathcal{L}^{CCDM}_{k} = \mathbb{E}_{\mathbf{y}_{0},\mathbf{x},k\sim \mathrm {U}[1,K]}(\mathcal{L}_{k}^{denoise}+\lambda \mathcal{L}_{k}^{contrast}).
\label{eq:6}
\vspace{-5pt}
\end{equation}

So far, we obtain the overall step-wise training procedure for CCDM, which is a $\lambda$-weighted combination of the vanilla denoising term in Eq. \ref{eq:1} and auxiliary contrastive item in Eq. \ref{eq:4}. The whole training algorithm is clarified in Appendix \ref{train_algo_sup}, which is efficient, end-to-end and seamlessly coupled with original simplified denoising diffusion.

\textbf{Theoretical insights.} Beyond the concrete method described above, we offer two-fold interpretations on how conditional diffusion forecasters can gain from auxiliary temporal contrastive learning. From the \textit{neural mutual information perspective}, Eq. \ref{eq:1} and Eq. \ref{eq:4} train the parameterized conditional denoiser $\bm{\epsilon}_{\theta}(\cdot)$ to simultaneously optimize two complementary lower bounds (i.e. predictive and contrastive) of the prediction-related mutual information $I_{\theta}(\mathbf{y}_{0}; \mathbf{x})$ between future forecasts and past conditioning time series. According to \citep{tsai2020self}, this composite learning method can enhance the representation efficiency to distill task-related information from accessible conditions. The contrastive scheme assists $\bm{\epsilon}_{\theta}(\cdot)$ to learn helpful negative instances, which can gain useful discriminative temporal patterns for accurate multivariate predictive distribution recovery and mitigate over-fitting on historical training time series. From the \textit{distribution generalization perspective}, explicitly optimizing the probabilities of unexpected negative samples can render $\bm{\epsilon}_{\theta}(\cdot)$ see more OOD regions that pure denoising fashion on positive in-distribution samples do not encompass. In time series forecasting, there always exists distribution shift between unforeseen testing data and historical training data. The contrastive term in Eq. \ref{eq:4} intuitively minimizes the possibility $\log p_{\theta}(\mathbf{y}_{0}^{(n)})$ of undesirable spurious forecasts by directly impeding $\bm{\epsilon}_{\theta}(\cdot)$ from correctly removing the noise over negative $\mathbf{y}_{0}^{(n)}$. This contrastive enforcement helps $\bm{\epsilon}_{\theta}(\cdot)$ avoid low-density areas and undergo more OOD areas during in-distribution training. In light of the arguments in \citep{wu2024your}, promoting the denoiser robustness in OOD regions in testing stage is crucial to sample plausible forecasts. 

Moreover, we reveal the upper bound of forecasting error on testing data for conditional diffusion models in Proposition 1. It obviously reflects that the efficacy of conditional diffusion forecasters is inextricably intertwined with the step-wise noise regression accuracy of trained $\bm{\epsilon}_{\theta}(\cdot)$ on unknown test time series. In this regard, resorting to temporal contrastive refinement or other auxiliary training regimes is sensible to boost conditional denoiser behaviors and final prediction outcomes.

\textbf{Proposition 1.} \textit{Let $q^{te}(\mathbf{y}_{0}|\mathbf{x})$ be the ground truth distribution of test time series, and $p^{te}_{\theta}(\mathbf{y}_{0}|\mathbf{x})$ be the approximated predictive distribution by the developed conditional diffusion model. Let the KL-divergence between $q^{te}(\mathbf{y}_{0}|\mathbf{x})$ and $p^{te}_{\theta}(\mathbf{y}_{0}|\mathbf{x})$ represent the resulting probabilistic forecasting error. Then the denoising diffusion-induced forecasting error is upper-bounded:}
\begin{equation}
\centering
\mathcal{D}_{KL}\left[q^{te}(\mathbf{y}_{0}|\mathbf{x})||p^{te}_{\theta}(\mathbf{y}_{0}|\mathbf{x})\right] \le \mathbb{E}_{\mathbf{x},\mathbf{y}_{0},\bm{\epsilon}_{k},k}\left[A_{k}\left\|\bm{\epsilon}_{\theta}\left(\sqrt{\bar{\alpha}_{k}}\mathbf{y}_{0}+\sqrt{1-\bar{\alpha}_{k}}\bm{\epsilon}_{k}, \mathbf{x}, k\right)-\bm{\epsilon}_{k}\right\|^{2}_{2}\right]+C.
\label{eq:7}
\end{equation}
\textit{Such upper bound is determined by the denoising behaviors of learned $\bm{\epsilon}_{\theta}(\cdot)$ on unknown test time series. $A_{k}$ is a step-wise constant related to noise schedule, and $C$ is a constant depending on test data quantities.} See Appendix \ref{proposition_proof} for the proof.

%% file: experiments.tex

\subsection{Experimental setup}
\textbf{Datasets.} We choose six multivariate time series datasets, i.e. \texttt{ETTh1}, \texttt{Exchange}, \texttt{Weather}, \texttt{Appliance}, \texttt{Electricity}, \texttt{Traffic}, which cover a wide range of temporal dynamics and channel number $D$ to completely gauge the probabilistic forecasting performance. We manually establish a more comprehensive benchmark with diverse values of lookback window $L$ and prediction horizon $H$, distinct from previous models which merely attest their generative forecasting capacity on a single short-term setup. Refer to Appendix. \ref{sec:dataset} for more details on datasets.

\textbf{Evaluation metrics.} We adopt two standard metrics to assess the quality of both probabilistic and deterministic forecasts resulting from the generated prediction intervals. CRPS (Continuous Ranked Probability Score) is used to assess the reliability of the estimated predictive distribution, and MSE (Mean Squared Error) is used to quantify the accuracy of calculated point forecasts. See Appendix \ref{eval_metric} for more details on metrics.

\textbf{Baselines.} We select five currently remarkable denoising diffusion-based generative forecasters for comparisons, including TimeGrad \citep{rasul2021autoregressive}, CSDI \citep{tashiro2021csdi}, SSSD \citep{alcaraz2022diffusion}, TimeDiff \citep{shen2023non}, TMDM \citep{li2023transformer}. Since these models do not shed light on outcomes on long-term probabilistic forecasting scenarios, we fully reproduce them on the newly constructed benchmark.

\textbf{Implementation details.} We normally execute the end-to-end contrastive diffusion training in Eq. \ref{eq:6} using 200 epochs. To reduce the contrastive learning costs on those cases which consume enormous computational resources, we also employ a cost-efficient two-stage training strategy. Concretely, we firstly pretrain a low-cost naive diffusion forecaster by Eq. \ref{eq:1} and fine-tune it by the total contrastive manner in Eq. \ref{eq:6} with only 30 epochs. We keep the temperature coefficient $\tau=0.1$ and randomly generate $S=100$ mulivariate profiles to compose prediction intervals. See Appendix \ref{sec:config} for more details on network architecture and contrastive training configurations in different forecasting setups. All experiments are conducted on a single NVIDIA A100 40GB GPU.

\subsection{Overall results}
\input{overall_table}

\begin{figure}[!t]
\vspace{-15pt}
\centering
\includegraphics[width=1.0\textwidth]{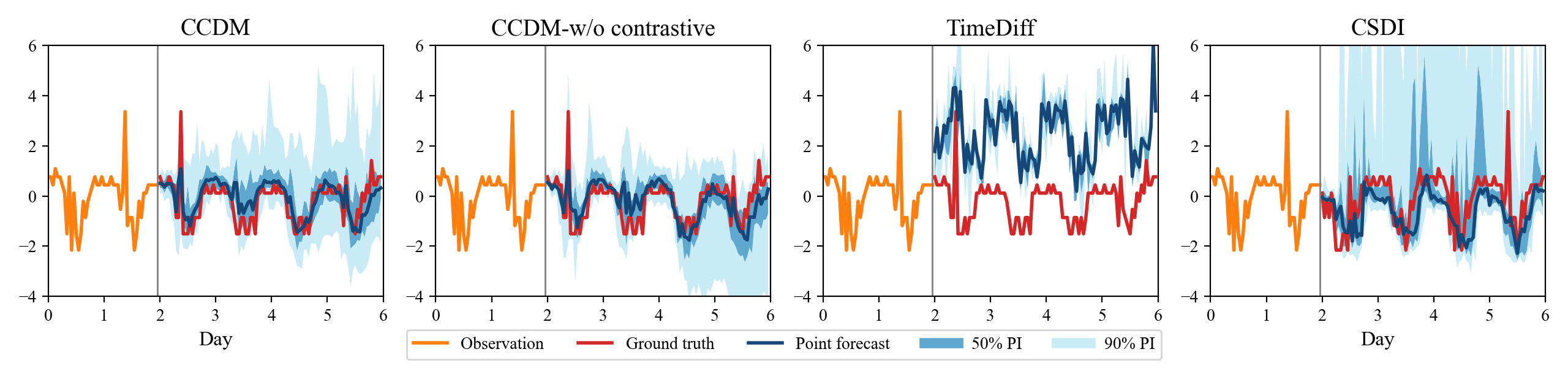}
\vspace{-10pt}
\caption{Comparison of generated point forecasts and prediction intervals on an Electricity channel.}
\vspace{-10pt}
\label{compare_forecasts}
\end{figure}

We demonstrate the devised CCDM model can outperform existing diffusion forecasters on most of the generative forecasting cases in Table \ref{overall_results}. Concretely, CCDM can attain the best outcomes on $16/24$ deterministic and $20/24$ probabilistic evaluations, with $9.10\%$ and $15.66\%$ average improvement of MSE and CRPS on these cases. Especially on two most difficult datasets \texttt{Electricity} and \texttt{Traffic}, CCDM garners notable progress of $8.94\%$, $10.59\%$ on MSE and $19.73\%$, $19.10\%$ on CRPS. These prominent increases reflect the devised channel-centric structure and contrastive refinement on the diffusion forecaster can enhance its representation efficiency of implicit predictive information on diverse prediction scenarios. The second-best model CSDI also manifests excellent forecasting ability especially on \texttt{Weather}, which has complex multivariate temporal correlations. The hybrid attention module in CSDI can well capture these relations but it entices high computational overhead and over-fitting to other datasets. TMDM and TimeDiff also attain small MSE on few cases due to their extra deterministic pre-training operations on conditioning encoders. Note that we completely replicate TimeGrad on the whole benchmark for the first time even with severe inference costs, and validate it can actually realize reasonable forecasting results. In Fig. \ref{compare_forecasts}, we depict different diffusion produced prediction intervals on one case. We can clearly see that CCDM's interval is much more faithful, while TimeDiff's area is sharper but loses diversity and accuracy. See Appendix \ref{pi_show} for more forecasting result showcases and Appendix \ref{time_analysis} on time cost analysis.

\subsection{Ablation study}
\label{sec:ablation_study}
To investigate  respective effects of each component, we remove the proposed denoising-based contrastive learning and channel-wise DiT structure, and exhibit the average metric degradation over different prediction horizons in Table \ref{ablation_table_short}. Without auxiliary contrastive diffusion training, we observe a mean performance drop of $8.33\%$ and $5.81\%$ on MSE and CRPS over the whole benchmark. This notable decrease indicates that the dedicated denoising-based contrastive refinement can enhance the utilization efficiency of conditional temporal predictive information and yield a more genuine multivariate predictive distribution. Due to the restriction of computational costs, such contrastive gains on \texttt{Electricity} and \texttt{Traffic} datasets are relatively smaller. We can amplify contrastive benefits on large-scale datasets by increasing the batch size and negative number within an iteration in the future. Regarding the influence of composite channel-aware management in conditional denoiser, we replace the channel-wise DiT modules by the same depth of linear dense encoders and incur a full channel-independence architecture. The average reduction on MSE and CRPS over the whole test settings are $19.80\%$ and $26.10\%$. This considerable drop reveals that the channel-mixing attention can empower the denoising network to integrate useful cross-variate temporal features in past observations and corrupted targets. Besides, the elevation degree induced by channel-centric DiT is consistent with the true variate correlations in real-world datasets. For instance, the performance decrease is less salient on \texttt{Electricity} dataset where the electricity consumption of different customers is not highly related to each other. Whilst on \texttt{ETTh1} and \texttt{Weather} datasets whose sensory measurements are heavily inter-correlated, the channel-mixing DiT can improve the diffusion forecasting capacity more vastly.
\input{ablation_table_short}

\begin{figure}[h]
\centering
\includegraphics[width=0.98\textwidth]{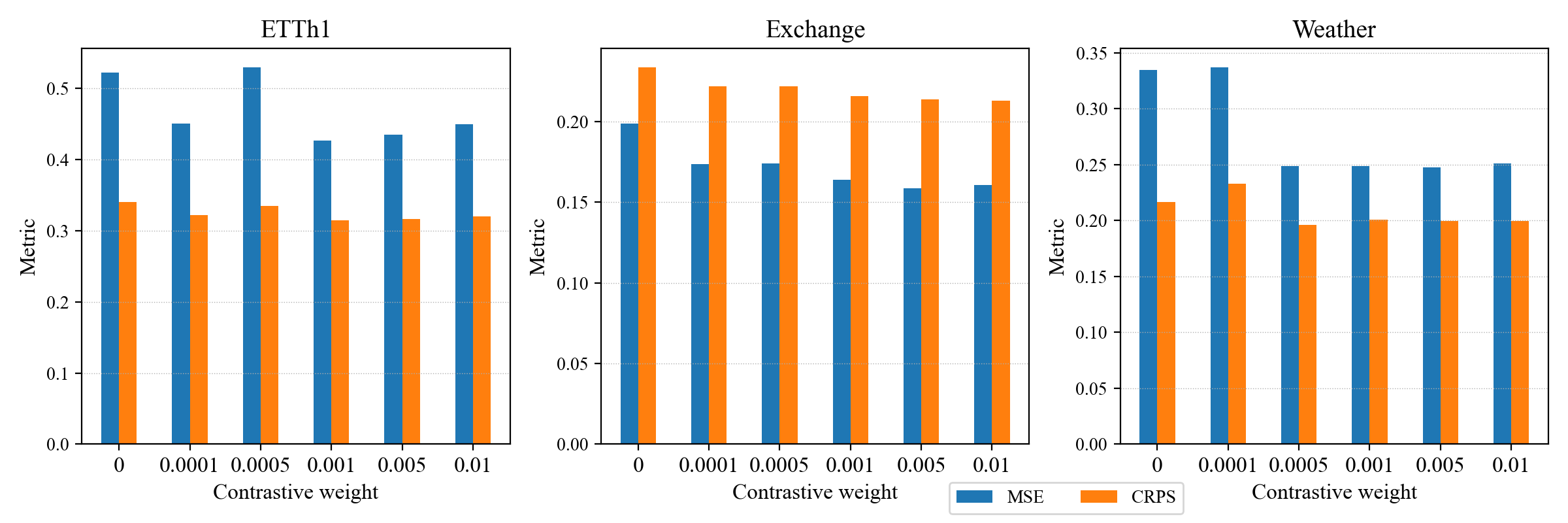}
\vspace{-10pt}
\caption{Forecasting results by varying contrastive weight $\lambda$ on three datasets with $H=168$.}
\vspace{-5pt}
\label{contrast_weight}
\end{figure}

\subsection{Contrastive refinement analysis}
Below, we empirically investigate the efficacy of the devised denoising-based temporal contrastive refinement, including three vital factors for contrastive learning practice and its generality on other existing diffusion forecasters.

\textbf{Influence of contrastive weight $\lambda$.} The complementary step-wise denoising-based contrastive loss in \ref{eq:4} can enhance the alignment between diffusion generated forecasts and given temporal predictive information. To elucidate the impact of contrastive refinement in different degrees on original diffusion optimization, we escalate the contrastive weight $\lambda$ in Eq. \ref{eq:5} from 0.0001 to 0.01 and display corresponding forecasting outcomes in Fig. \ref{contrast_weight}. We can totally find that imposing contrastive regime on denoising diffusion training can indeed promote the generative forecasting capacity, and the gain margin moderately fluctuates among various weights and datasets. Roughly, a modest weight between 0.0005 and 0.005 can lead to better improvement. We also empirically observe that larger $\lambda$ can accelerate the diffusion training convergence. See Appendix \ref{contrast_analysis} for more detailed analysis on the influence of negative number $N$ and temperature $\tau$.

\input{transfer_table}
\textbf{Generality of contrastive training.} We add the step-wise denoising contrastive training presented in \ref{eq:4} to two existing diffusion forecasters to validate its generality on conditional time series diffusion learning. From the results shown in Table \ref{transfer_table}, it is obvious that CSDI's generative forecasting ability can be further enhanced by contrastive diffusion training. Its hybrid attention network can represent complex temporal patterns more properly by handling more OOD negative samples. While for TimeDiff which owns extra pre-trained auto-regressive conditioning encoders, CRPS values constantly decrease but some unexpected increases appear on MSE. It may stem from the side effect of redundant contrastive procedure conveyed to the well-behaved deterministic pre-training strategy.

%% file: overall_table.tex
\begin{table}[!t]
\vspace{-15pt}
\centering
\caption{Overall comparisons w.r.t MSE and CRPS on six real-world datasets with diverse horizon $H \in \left \{ 96,168,336,720 \right \} $. The best and second-best results are boldfaced and underlined.}
\label{overall_results}
\resizebox{1.0\textwidth}{!}{
\begin{tabular}{cc|cc|cc|cc|cc|cc|cc}
\toprule
\multicolumn{2}{c|}{Methods}           & \multicolumn{2}{c|}{CCDM} & \multicolumn{2}{c|}{TMDM}        & \multicolumn{2}{c|}{TimeDiff}    & \multicolumn{2}{c|}{SSSD}        & \multicolumn{2}{c|}{CSDI}         & \multicolumn{2}{c}{TimeGrad}    \\
\midrule
\multicolumn{2}{c|}{Metrics}                                                 & MSE             & CRPS            & MSE             & CRPS         & MSE             & CRPS         & MSE          & CRPS         & MSE             & CRPS            & MSE        & CRPS            \\ \midrule
\multicolumn{1}{c|}{\multirow{5}{*}{\rotatebox[origin=c]{90}{ETTh1}}}       & 96                       & \textbf{0.3856} & \textbf{0.2935} & 0.4692          & 0.3952       & {\ul 0.4025}    & {\ul 0.3942} & 1.0984       & 0.5622       & 1.1013          & 0.5794          & 1.1730     & 0.6223          \\
\multicolumn{1}{c|}{}                             & 168                      & \textbf{0.4267} & \textbf{0.3142} & 0.5296          & 0.4163       & {\ul 0.4397}    & 0.4170       & 0.6067       & {\ul 0.4046} & 1.1013          & 0.5794          & 1.1554     & 0.5970          \\
\multicolumn{1}{c|}{}                             & 336                      & {\ul 0.5312}    & \textbf{0.3452} & 0.5862          & 0.4655       & \textbf{0.4943} & {\ul 0.4488} & 0.9330       & 0.5421       & 1.0459          & 0.6223          & 1.1403     & 0.5883          \\
\multicolumn{1}{c|}{}                             & 720                      & \textbf{0.5642}    & \textbf{0.4870} & 0.7083          & 0.5335       & {\ul 0.5779} & {\ul 0.5145} & 1.3776       & 0.7035       & 1.0081          & 0.5952          & 1.2529     & 0.6498          \\ \cmidrule{2-14} 
\multicolumn{1}{c|}{}                             & \multicolumn{1}{l|}{Avg} & \textbf{0.4769}    & \textbf{0.3600} & 0.5733          & 0.4526       & {\ul 0.4786} & {\ul 0.4418} & 1.0039       & 0.5531       & 1.0642          & 0.5941          & 1.1804     & 0.6144          \\ \midrule
\multicolumn{1}{c|}{\multirow{5}{*}{\rotatebox[origin=c]{90}{Exchange}}}    & 96                       & \textbf{0.0905} & \textbf{0.1545} & 0.1278          & {\ul 0.2112} & {\ul 0.1106}    & 0.2349       & 0.5551       & 0.4569       & 0.2551          & 0.2901          & 1.8655     & 1.0439          \\
\multicolumn{1}{c|}{}                             & 168                      & \textbf{0.1638} & \textbf{0.2159} & 0.2791          & 0.3210       & {\ul 0.2050}    & {\ul 0.3187} & 0.4517       & 0.3602       & 0.8050          & 0.5093          & 1.1638     & 0.8374          \\
\multicolumn{1}{c|}{}                             & 336                      & \textbf{0.4407} & \textbf{0.3517} & {\ul 0.4572}    & 0.4426       & 0.5834          & 0.5472       & 0.5641       & {\ul 0.4106} & 0.6179          & 0.4786          & 1.9264     & 1.0465          \\
\multicolumn{1}{c|}{}                             & 720                      & {\ul 1.1685}    & \textbf{0.5864} & 2.5625          & 1.0828       & \textbf{0.9096} & 0.7128       & 1.3686       & {\ul 0.6386} & 1.3816          & 0.7423          & 2.4034     & 1.1478          \\ \cmidrule{2-14} 
\multicolumn{1}{c|}{}                             & \multicolumn{1}{l|}{Avg} & {\ul 0.4659}    & \textbf{0.3271} & 0.8567          & 0.5144       & \textbf{0.4522} & {\ul 0.4534} & 0.7349       & 0.4666       & 0.7649          & 0.5051          & 1.8398     & 1.0189          \\ \midrule
\multicolumn{1}{c|}{\multirow{5}{*}{\rotatebox[origin=c]{90}{Weather}}}     & 96                       & {\ul 0.2669}    & \textbf{0.1904} & 0.2768          & 0.2273       & 0.3842          & 0.3441       & 0.6103       & 0.3878       & \textbf{0.2608} & {\ul 0.2127}    & 0.5628     & 0.3445          \\
\multicolumn{1}{c|}{}                             & 168                      & \textbf{0.2489} & \textbf{0.2006} & 0.2864          & 0.2519       & 0.3566          & 0.3192       & {\ul 0.2796} & {\ul 0.2060} & 0.2930          & 0.2286          & 0.4141     & 0.2880          \\
\multicolumn{1}{c|}{}                             & 336                      & \textbf{0.2870} & {\ul 0.2258}    & 0.3494          & 0.3007       & 0.4805          & 0.3591       & 0.3189       & 0.2355       & {\ul 0.2918}    & \textbf{0.2193} & 0.5462     & 0.3549          \\
\multicolumn{1}{c|}{}                             & 720                      & 0.5627          & 0.4083          & {\ul 0.3975}    & 0.3365       & 0.5052          & 0.3880       & 0.6880       & 0.4179       & \textbf{0.3803} & \textbf{0.2770} & 0.4774     & {\ul 0.3221}    \\ \cmidrule{2-14} 
\multicolumn{1}{c|}{}                             & \multicolumn{1}{l|}{Avg} & 0.3414          & {\ul 0.2563}    & {\ul 0.3275}    & 0.2791       & 0.4316          & 0.3526       & 0.4742       & 0.3118       & \textbf{0.3065} & \textbf{0.2344} & 0.5001     & 0.3274          \\ \midrule
\multicolumn{1}{c|}{\multirow{5}{*}{\rotatebox[origin=c]{90}{Appliance}}}   & 96                       & 0.6988          & \textbf{0.4138} & {\ul 0.6858}    & 0.4678       & 0.7328          & 0.5740       & 1.1954       & 0.6504       & \textbf{0.6823} & {\ul 0.4334}    & 1.6748     & 0.8397          \\
\multicolumn{1}{c|}{}                             & 168                      & \textbf{0.6266} & \textbf{0.4020} & 0.7153          & 0.5232       & {\ul 0.6468}    & 0.5562       & 0.7841       & 0.4776       & 0.7176          & {\ul 0.4560}    & 1.8901     & 0.8858          \\
\multicolumn{1}{c|}{}                             & 336                      & \textbf{0.9119} & \textbf{0.5036} & 1.0310          & 0.6590       & {\ul 0.9531}    & 0.6822       & 1.8822       & 0.8002       & 1.0565          & {\ul 0.5675}    & 1.8506     & 0.8661          \\
\multicolumn{1}{c|}{}                             & 720                      & 1.5798          & 0.8620          & \textbf{1.3937} & {\ul 0.8272} & {\ul 1.4327}    & 0.8809       & 3.3226       & 1.1225       & 1.7347          & \textbf{0.7982} & 2.4393     & 1.0083          \\ \cmidrule{2-14} 
\multicolumn{1}{c|}{}                             & \multicolumn{1}{l|}{Avg} & {\ul 0.9543}    & \textbf{0.5454} & 0.9565          & 0.6193       & \textbf{0.9414} & 0.6733       & 1.7961       & 0.7627       & 1.0478          & {\ul 0.5638}    & 1.9637     & 0.9000          \\ \midrule
\multicolumn{1}{c|}{\multirow{5}{*}{\rotatebox[origin=c]{90}{Electricity}}} & 96                       & 0.2102          & \textbf{0.2182} & \textbf{0.1954} & 0.3113       & {\ul 0.1960}    & 0.3123       & 0.2444       & {\ul 0.2346} & 0.2560          & 0.2571          & 0.3733     & 0.3259          \\
\multicolumn{1}{c|}{}                             & 168                      & \textbf{0.1678} & {\ul 0.2014}    & 0.1908          & 0.3037       & 0.1907          & 0.3043       & 0.2001       & 0.2249       & {\ul 0.1754}    & \textbf{0.1985} & 0.3676     & 0.3083          \\
\multicolumn{1}{c|}{}                             & 336                      & \textbf{0.1683} & \textbf{0.2014} & 0.2042          & 0.3165       & 0.2047          & 0.3172       & 0.1941       & 0.2245       & {\ul 0.1803}    & {\ul 0.2043}    & 0.4249     & 0.3497          \\
\multicolumn{1}{c|}{}                             & 720                      & \textbf{0.1994} & \textbf{0.2232} & 0.2282          & 0.3338       & {\ul 0.2277}    & {\ul 0.3336} & 0.3743       & 0.3680       & 0.9932          & 0.5678          & 0.4299     & 0.3479          \\ \cmidrule{2-14} 
\multicolumn{1}{c|}{}                             & \multicolumn{1}{l|}{Avg} & \textbf{0.1864} & \textbf{0.2111} & {\ul 0.2047}    & 0.3163       & 0.2048          & 0.3169       & 0.2532       & {\ul 0.2630} & 0.4012          & 0.3069          & 0.3989     & 0.3330          \\ \midrule
\multicolumn{1}{c|}{\multirow{5}{*}{\rotatebox[origin=c]{90}{Traffic}}}     & 96                       & 1.0282          & \textbf{0.4093} & {\ul 0.9692}    & 0.5894       & \textbf{0.9684} & 0.5859       & 1.0363       & 0.4445       & 1.1154          & {\ul 0.4240}    & 1.2259     & 0.4667          \\
\multicolumn{1}{c|}{}                             & 168                      & \textbf{0.6881} & \textbf{0.3077} & 0.8632          & 0.5254       & {\ul 0.8553}    & 0.5192       & 0.9551       & {\ul 0.4289} & 1.6000          & 0.6701          & 1.3282     & 0.5510          \\
\multicolumn{1}{c|}{}                             & 336                      & \textbf{0.6863} & \textbf{0.3358} & 0.8874          & 0.5562       & {\ul 0.8834}    & 0.5538       & 0.9283       & 0.5140       & 1.5724          & 0.6780          & 1.0447     & {\ul 0.3817}    \\
\multicolumn{1}{c|}{}                             & 720                      & \textbf{0.9357} & \textbf{0.4519} & {\ul 1.0258}    & 0.6383       & 1.0270          & 0.6387       & 1.0635       & 0.5515       & 1.5428          & 0.6696          & 1.1753     & {\ul 0.4604}    \\ \cmidrule{2-14} 
\multicolumn{1}{c|}{}                             & \multicolumn{1}{l|}{Avg} & \textbf{0.8346} & \textbf{0.3762} & 0.9364          & 0.5773       & {\ul 0.9335}    & 0.5744       & 0.9958       & 0.4847       & 1.4577          & 0.6104          & 1.1935     & {\ul 0.4650}    
\\ \bottomrule
\end{tabular}
}
\vspace{-5pt}
\end{table}

%% file: ablation_table_short.tex
\begin{table}[ht]
\centering
\caption{Average MSE and CRPS degradation resulting from the ablation of denoising-based contrastive learning or channel-wise DiT module. Full results can be found in Appendix \ref{sec:full_ablation}.}
\label{ablation_table_short}
\resizebox{1.0\textwidth}{!}{
\begin{tabular}{c|cccc|cccc}
\toprule
Models      & \multicolumn{4}{c|}{w/o contrastive refinement}                  & \multicolumn{4}{c}{w/o channel-wise DiT}                         \\ \hline
Metrics     & MSE    & \multicolumn{1}{c|}{Degradation} & CRPS   & Degradation & MSE    & \multicolumn{1}{c|}{Degradation} & CRPS   & Degradation \\ \hline
ETTh1       & 0.5508 & \multicolumn{1}{c|}{15.97\%}     & 0.3889 & 8.55\%      & 0.5956 & \multicolumn{1}{c|}{23.34\%}     & 0.5816 & 58.20\%     \\
Exchange    & 0.4966 & \multicolumn{1}{c|}{11.52\%}     & 0.3403 & 5.24\%      & 0.4924 & \multicolumn{1}{c|}{12.18\%}     & 0.3555 & 11.00\%     \\
Weather     & 0.3816 & \multicolumn{1}{c|}{13.00\%}     & 0.2695 & 4.77\%      & 0.4843 & \multicolumn{1}{c|}{40.28\%}     & 0.3336 & 28.92\%     \\
Appliance   & 1.0220 & \multicolumn{1}{c|}{7.18\%}      & 0.5818 & 5.99\%      & 1.1183 & \multicolumn{1}{c|}{16.96\%}     & 0.7231 & 32.70\%     \\
Electricity & 0.1887 & \multicolumn{1}{c|}{1.20\%}      & 0.2144 & 1.57\%      & 0.1973 & \multicolumn{1}{c|}{5.71\%}      & 0.2137 & 1.25\%      \\
Traffic     & 0.8439 & \multicolumn{1}{c|}{1.08\%}      & 0.4130 & 8.73\%      & 1.0084 & \multicolumn{1}{c|}{20.30\%}     & 0.4675 & 24.50\%     \\ 
\bottomrule
\end{tabular}
}
\vspace{-10pt}
\end{table}

%% file: transfer_table.tex
\begin{table}[!t]
\vspace{-20pt}
\centering
\caption{Forecasting performance promotion induced by applying denoising-based contrastive training to two existing conditional diffusion forecasters.}
\label{transfer_table}
\resizebox{0.9\textwidth}{!}{
\begin{tabular}{cc|cccc|cccc}
\toprule
\multicolumn{2}{c|}{Methods}                         & \multicolumn{4}{c|}{TimeDiff}                                & \multicolumn{4}{c}{CSDI}                                     \\ \midrule
\multicolumn{2}{c|}{Metrics}                         & MSE    & \multicolumn{1}{c|}{Promotion} & CRPS   & Promotion & MSE    & \multicolumn{1}{c|}{Promotion} & CRPS   & Promotion \\ \midrule
\multicolumn{1}{c|}{\multirow{4}{*}{\rotatebox[origin=c]{90}{ETTh1}}}    & 96  & 0.4143 & \multicolumn{1}{c|}{-2.93\%}   & 0.3491 & 11.44\%   & 0.6559 & \multicolumn{1}{c|}{40.44\%}   & 0.4371 & 24.56\%   \\
\multicolumn{1}{c|}{}                          & 168 & 0.4715 & \multicolumn{1}{c|}{-7.23\%}   & 0.3753 & 10.00\%   & 0.5894 & \multicolumn{1}{c|}{29.53\%}   & 0.3851 & 25.76\%   \\
\multicolumn{1}{c|}{}                          & 336 & 0.5073 & \multicolumn{1}{c|}{-2.63\%}   & 0.4025 & 10.32\%   & 0.9920 & \multicolumn{1}{c|}{5.15\%}    & 0.5644 & 9.30\%    \\
\multicolumn{1}{c|}{}                          & 720 & 0.5291 & \multicolumn{1}{c|}{6.19\%}    & 0.4338 & 14.44\%   & 0.7744 & \multicolumn{1}{c|}{23.18\%}   & 0.7010 & -17.78\%  \\ \midrule
\multicolumn{1}{c|}{\multirow{4}{*}{\rotatebox[origin=c]{90}{Exchange}}} & 96  & 0.0901 & \multicolumn{1}{c|}{18.54\%}   & 0.1722 & 26.69\%   & 0.1589 & \multicolumn{1}{c|}{37.71\%}   & 0.2082 & 28.23\%   \\
\multicolumn{1}{c|}{}                          & 168 & 0.1588 & \multicolumn{1}{c|}{22.54\%}   & 0.2312 & 27.46\%   & 0.4096 & \multicolumn{1}{c|}{49.12\%}   & 0.3840 & 24.60\%   \\
\multicolumn{1}{c|}{}                          & 336 & 0.6345 & \multicolumn{1}{c|}{-8.76\%}   & 0.4293 & 21.55\%   & 0.5664 & \multicolumn{1}{c|}{8.33\%}    & 0.4110 & 14.12\%   \\
\multicolumn{1}{c|}{}                          & 720 & 0.9735 & \multicolumn{1}{c|}{-7.03\%}   & 0.6941 & 2.62\%    & 1.3642 & \multicolumn{1}{c|}{1.26\%}    & 0.6392 & 13.89\%   \\ 
\bottomrule
\end{tabular}
}
\vspace{-10pt}
\end{table}

%% file: related_work_short.tex
\textbf{Channel-oriented multivariate forecasting.} Recent progress on multivariate deterministic prediction \citep{liu2023itransformer, lu2023arm, chen2024similarity, han2024softs} indicate that learning channel-centric temporal properties (including single-channel dynamics and cross-channel correlations) is of significant importance. Both channel-independent and channel-fusing time series processing are crucial to improve the forecasting performance. But the effectiveness of such channel manipulation structures is rarely investigated in diffusion-based multivariate probabilistic forecasting, where the extra influence of imposed channel noise in varying degrees should also be addressed. To tackle this barrier, we blend both channel-independent and channel-mixing modules in the conditional diffusion denoiser to boost its forecasting ability on multivariate cases. 

\textbf{Time series diffusion models.} Diffusion models have been actively applied to tackle a wide scope of time series tasks, including synthesis \citep{yuan2024diffusion, narasimhan2024time}, forecasting \citep{rasul2021autoregressive}, imputation \citep{tashiro2021csdi} and anomaly detection \citep{chen2023imdiffusion}. Their common goal is to derive a high-quality conditional temporal distribution aligned with diverse input contexts, such as statistical properties in constrained generation \citep{coletta2024constrained} and historical records. A valid solution is to inject useful temporal properties into iterative diffusion learning \citep{yuan2024diffusion, bilovs2023modeling} or to develop gradient-based guidance schemes \citep{coletta2024constrained}. But there are still rooms to enhance them from the aspect of training methods and denoiser architectures. To bridge this gap for multivariate forecasting, we exclusively design a channel-aware denoiser and explicitly enhance the predictive mutual information between past observations and future forecasts by an adapted temporal contrastive diffusion learning. Even though several works have applied contrastive diffusion to cross-modal content creation \citep{wang2024multi, zhu2022discrete}, its efficacy on time series generative modeling have not yet been well explored. And reasonable interpretations on such contrastive diffusion merits are also scanty. See Appendix \ref{related_work} for more detailed related work, which also covers universal temporal contrastive learning.

%% file: conclusion.tex

In this work, we propose the channel-aware contrastive conditional diffusion model named CCDM for probabilistic  forecasts on multivariate time series. CCDM can capture intrinsic prediction-related temporal information hidden in observed conditioning time series using an efficient channel-centric denoiser architecture and information-maximizing denoising-based contrastive refinement. Extensive experiments demonstrate the exceptional forecasting capability of CCDM over existing time series diffusion models. In future work, we plan to reduce the training costs imposed by additional temporal contrastive learning, and extend this contrastive diffusion method to general time series analysis and other cross-domain synthesis tasks.

%% file: appendix.tex
\subsection{Proof for Proposition 1}
\label{proposition_proof}
\input{proof}

\subsection{Additional Discussions on related works}
\label{related_work}
\input{related_work}

\subsection{Negative time series augmentation}
\label{neg_aug}
\begin{figure}[!h]
\centering
\includegraphics[width=0.95\textwidth]{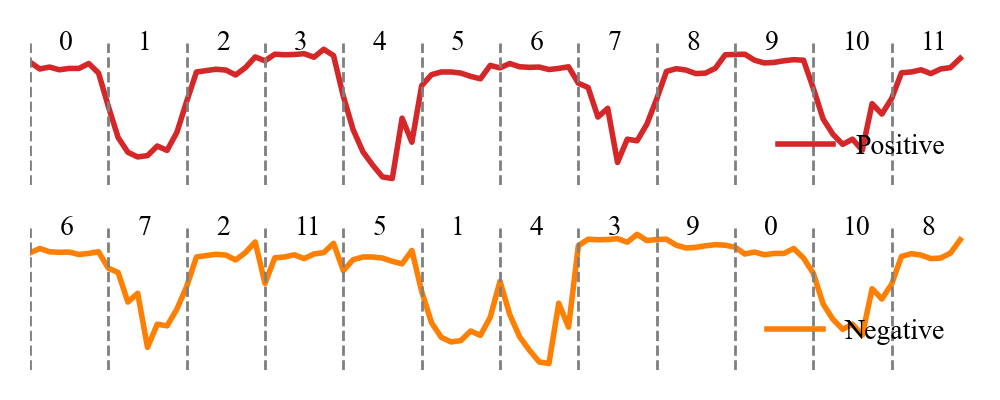}
\caption{One diagram of the variation-based time series augmentation method.}
\vspace{-10pt}
\label{neg_variation}
\end{figure}

To enable contrastive learning,  we employ two types of augmentation methods to produce negative multivariate time series $\mathbf{y}_{0}^{(n)}$. The first way is to alter the ground truth temporal variations of each univariate time series by patch shuffling, since recovering the correct temporal evolution is a vital challenge for time series diffusion models. As is shown in Fig. \ref{neg_variation}, we divide a given sequence into an array of sub-series patches and randomly shuffle their orders to change original temporal dynamics. The second way is to scale up or scale down the magnitudes of individual time points, as an ideal prediction interval should well cover every points without any of them falling outside. Thus, for each positive target $\mathbf{y}_{0}$, we uniformly sample a scaling factor $a_{d}$ between $\left [ 0, 0.5 \right ] \cup \left [ 1.5, 2.0 \right ]$ and impose it on each channel by $a_{d}\cdot \mathbf{y}_{0}^{d} \in \mathbb{R}^{H}$. We find both ways of generating negative samples would help the diffusion model learn more realistic time series samples. 

\subsection{Training algorithm}
\label{train_algo_sup}
We elucidate the step-wise denoising-based contrastive diffusion training algorithm in Algorithm \ref{train_algo}.
\input{train_algo}

\subsection{Dataset description}
\label{sec:dataset}
We present the dataset usage in Table \ref{dataset_table}, where the channel number $D$, sampling rate, train/val/test split size and own field are clarified. We also provide accessible repositories for these datasets below:

1) \texttt{ETTh1}: \href{https://github.com/zhouhaoyi/ETDataset}{https://github.com/zhouhaoyi/ETDataset} \\
2) \texttt{Exchange}: \href{https://github.com/laiguokun/multivariate-time-series-data}{https://github.com/laiguokun/multivariate-time-series-data} \\
3) \texttt{Weather}: \href{https://www.bgc-jena.mpg.de/wetter/}{https://www.bgc-jena.mpg.de/wetter/} \\
4) \texttt{Appliance}: \href{https://archive.ics.uci.edu/dataset/374/appliances+energy+prediction}{https://archive.ics.uci.edu/dataset/374/appliances+energy+prediction} \\
5) \texttt{Electricity}: \href{https://archive.ics.uci.edu/dataset/321/electricityloaddiagrams20112014}{https://archive.ics.uci.edu/dataset/321/electricityloaddiagrams20112014} \\
6) \texttt{Traffic}: \href{https://pems.dot.ca.gov/}{https://pems.dot.ca.gov/}
\input{dataset_table}

\subsection{Evaluation metrics}
\label{eval_metric}
To assess the accuracy and reliability of estimated multivariate predictive distribution, we adopt two common metrics, i.e. MSE and CRPS to quantify both deterministic and probabilistic forecasting performance of generated prediction intervals. Assume $\mathbf{y}_{0}$ is the ground-truth time series, $\{\hat{\mathbf{y}}_{0}^{(s)}\}_{s=1}^{S}$ is the produced prediction set, and let its $50\%$-quantile trajectory $\bar{\mathbf{y}}_{0}$ signify the point forecast, then two metrics can be calculated in a point-wise form over all channels and timestamps:
\begin{equation}
\centering
\label{eq:15}
MSE=\frac{1}{HD} \left \| \mathbf{y}_{0}-\bar{\mathbf{y}}_{0} \right \| _{2}^{2};
\end{equation}
\begin{equation}
\centering
\label{eq:16}
CRPS=\frac{1}{HD} \sum_{d=1}^{D} \sum_{t=1}^{H} \int_{R}^{} (F(\hat{y}_{td}) - \mathbb{I} \{y_{td} \le \hat{y}_{td}\})^{2}\mathrm{d}\hat{y}_{td};
\end{equation}
where $y_{td}$ indicates the $t$-th point value of the $d$-th univariate time series. $F$ is the empirical cumulative distribution function.

\subsection{Experimental configurations}
\label{sec:config}
In Table \ref{config_table}, we detail the conditional diffusion model configurations on different forecasting scenarios, including the channel-aware DiT compositions and diffusion noise scheduling. We simply preserve the layers of input and output channel-independent dense encoders identical to the depth of attention modules, i.e. $n_{att}=n_{enc}=n_{dec}=2$. One observation is that the designed channel-centric conditional denoising network can be easily scalable with diverse forecasting scenarios by merely adjusting the hidden representation dimension $e_{hid}$, which changes compatibly with the prediction horizon $H$.

In Table \ref{implement_table}, we shed light on the concrete contrastive training configurations for the main comparison outcomes presented in Table \ref{overall_results}. We adopt the two-stage separate training on Weather, Electricity and Traffic datasets to reduce the training time and memory consumption. The best contrastive weight is chosen from $\left \{ 0.001, 0.0005, 0.0001, 0.00005 \right \} $. Due to the GPU memory limitation, we have to turn down the negative sample number and batch size on Electricity and Traffic datasets with hundreds of channels, which could restrict the resulting final forecast performance. The initial learning rates are also displayed for the full reproduction on the newly adopted benchmark.

\input{config_table}
\input{implement_table}

\subsection{Computational time analysis}
\label{time_analysis}
We compare the both training and inference time costs of disparate diffusion forecasters in Table \ref{time_table}.
It is obvious that the auxiliary contrastive learning indeed aggravates the burden of vanilla denoising diffusion training for the sake of a higher quality of multivariate predictive distribution. Thus we adopt the two-stage separate strategy to accelerate the training process. The sequential generation procedure of our CCDM method is notably faster than other models, which indicates the designed channel-centric denoiser architecture can be efficiently scalable to diverse forecasting settings. Besides, the deterministic autoregressive pretraining in TimeDiff, hybrid attention layers in CSDI and point-wise amortized diffusion in TimeGrad can magnify their time consumption to different extents.
\input{time_table}

\subsection{Full results on ablation study}
\label{sec:full_ablation}
We illuminate the full forecasting outcomes corresponding to the ablation study of Section \ref{sec:ablation_study} in Table \ref{ablation_table}. In a nutshell, the performance promotion margins derived from such denoiser architecture and contrastive refinement innovations vary among different forecasting scenarios. It still requires careful settings on channel-aware denoising networks and auxiliary contrastive training to achieve the optimal results for a specific time series field and prediction setup.
\input{ablation_table}

\subsection{More analysis on contrastive refinement}
\label{contrast_analysis}
\begin{figure}[h]
\centering
\includegraphics[width=0.98\textwidth]{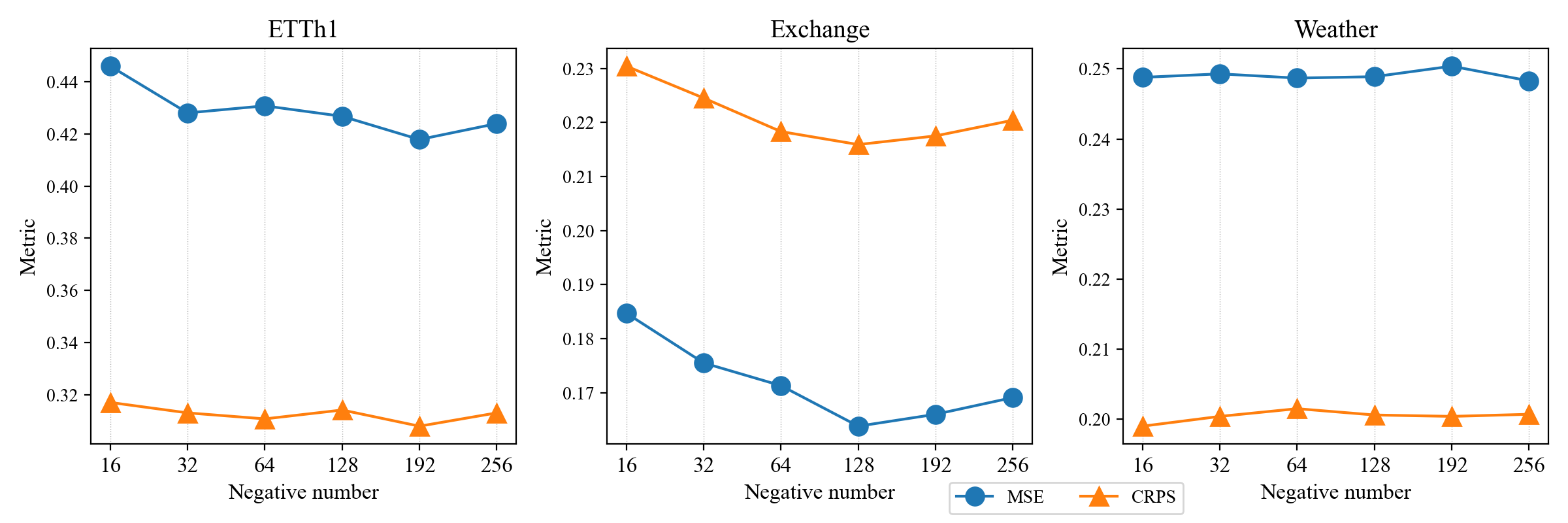}
\caption{Forecasting results by different numbers of negative samples.}
\vspace{-10pt}
\label{neg_num}
\end{figure}

\textbf{Influence of negative number $N$.} It is claimed in previous works on visual contrastive representation learning \cite{oord2018representation, chen2020simple} that a larger number of negative samples within a training iteration can bring out more informative latent features for downstream vision recognition tasks. To probe the influence of number of negative sample $N$ on the specialized contrastive time series diffusion model for multivariate forecasting, we change $N$ in the range from $16$ to $256$ and showcase pertaining outcomes in Fig. \ref{neg_num}. We can observe that the optimal $N$ is $192$, $128$ and $16$ on three datasets and two quantitative metrics of each dataset exhibit distinctive changing trends. This phenomenon suggests that the real impact of negative sample number on contrastive training gains is relatively intractable, which is not amenable to the law in visual contrastive self-supervised pretraining. It could also be caused by the substantially smaller amount of training corpus in time series than images. We should determine the best number of negative instances in light of concrete data characteristics along with other training hyper-parameters.

\textbf{Influence of temperature coefficient $\tau$.} The proposed denoising-based contrastive diffusion loss in Eq. \ref{eq:4} is in a canonical softmax form. According to the gradient analysis for the universal softmax-based contrastive loss in \cite{wang2021understanding}, the temperature $\tau$ is a critical factor to control the penalty magnitude on various negative samples. To attain the contrastive improvement on conditional denoiser training, maintaining $\tau$ within an appropriate interval is significant. We assign four values to $\tau$ and illustrate quantitative results in Table \ref{temp_table}. We can apparently observe that $[0.05, 0.1]$ could be a reasonable range on \texttt{ETTh1} and $[0.1, 0.5]$ is also valid for other two datasets.
\input{temp_table}

\subsection{More showcases on prediction intervals}
\label{pi_show}
In Fig. \ref{etth1_pi}-\ref{traffic_pi} below, we visualize more prediction intervals generated by the proposed CCDM on six datasets. The legend for each figure is identical to Fig. \ref{compare_forecasts}. For each task's result visualization, we just display the first 7 or 8 variates and present two random samples on the $L=48, H=96$ setting. 

\begin{figure}[ht]
\centering
\includegraphics[width=1.0\textwidth]{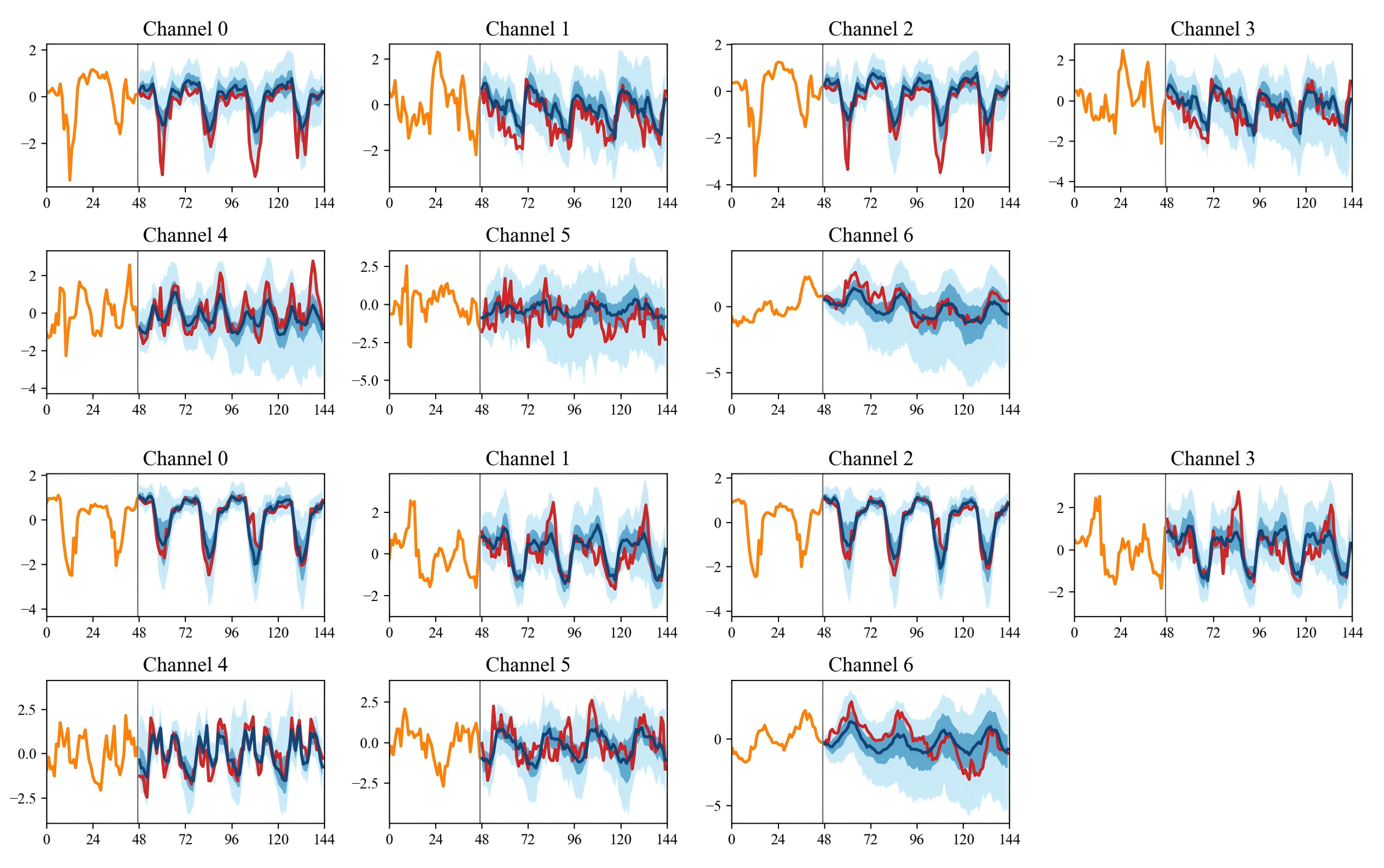}
\vspace{-10pt}
\caption{ETTh1 prediction intervals of total 7 channels.}
\vspace{-10pt}
\label{etth1_pi}
\end{figure}

\begin{figure}[ht]
\centering
\includegraphics[width=1.0\textwidth]{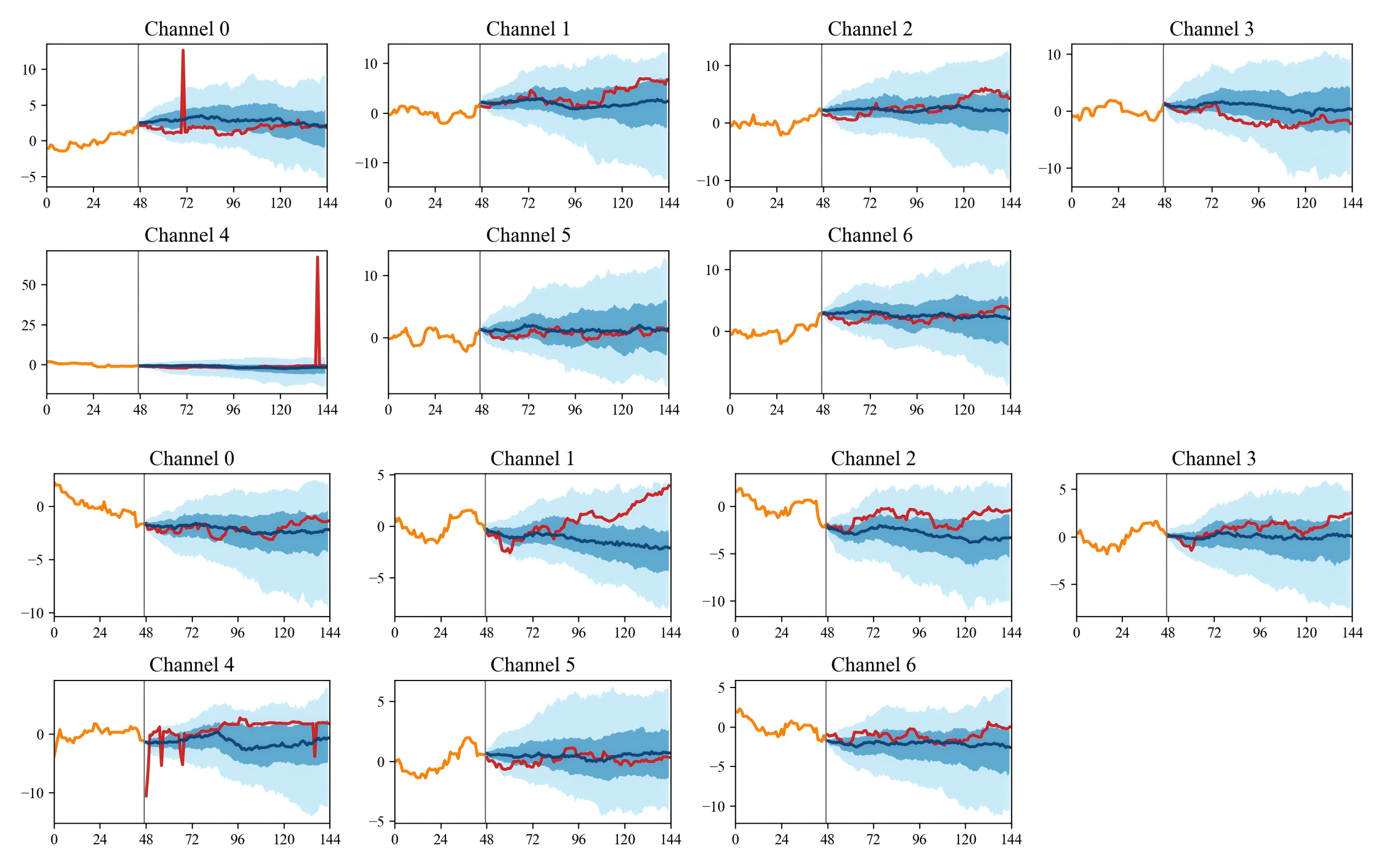}
\vspace{-10pt}
\caption{Exchange prediction intervals of total 7 channels.}
\vspace{-10pt}
\label{exchange_pi}
\end{figure}

\begin{figure}[ht]
\centering
\includegraphics[width=1.0\textwidth]{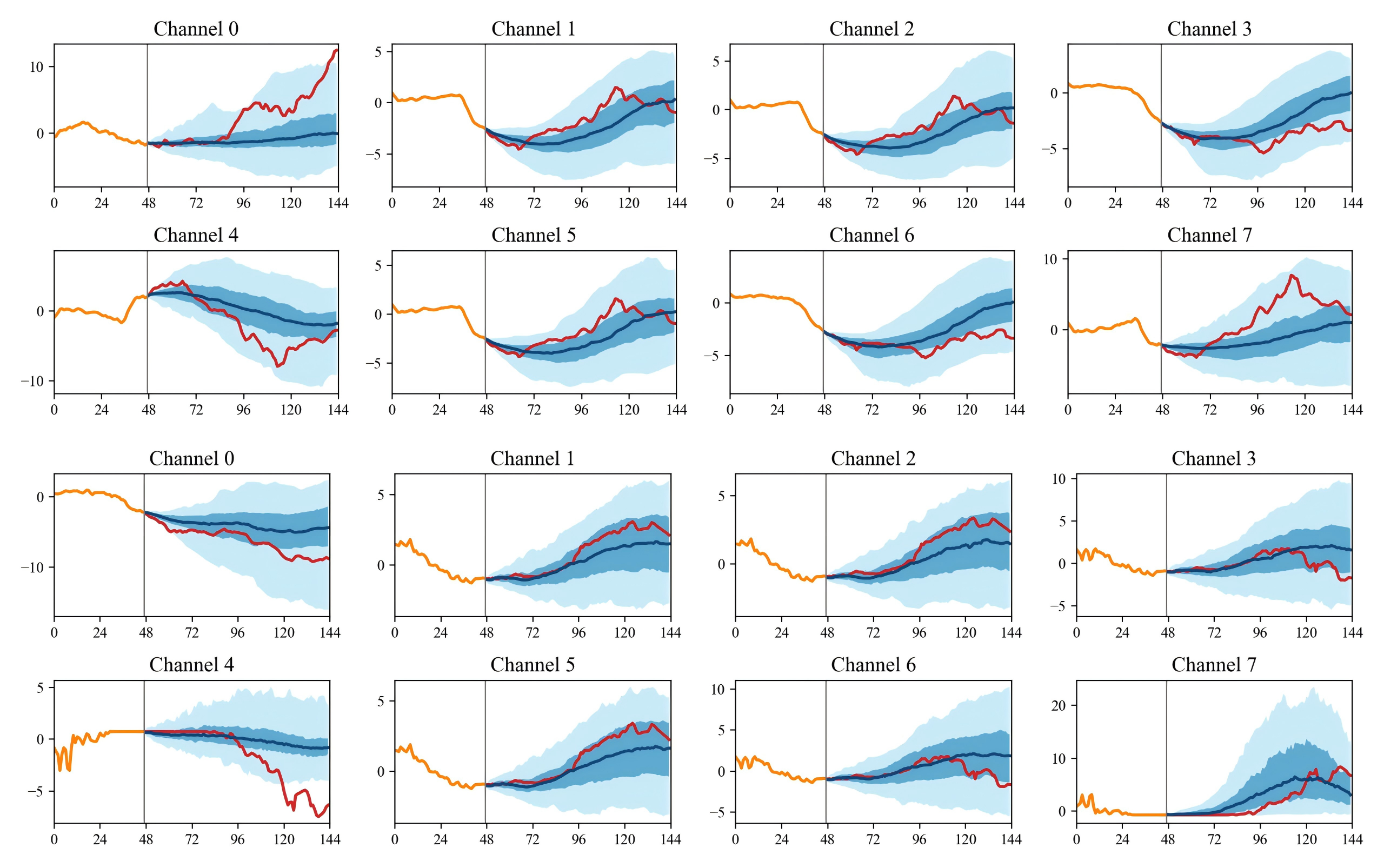}
\vspace{-10pt}
\caption{Weather prediction intervals of first 8 channels.}
\vspace{-10pt}
\label{weather_pi}
\end{figure}

\begin{figure}[ht]
\centering
\includegraphics[width=1.0\textwidth]{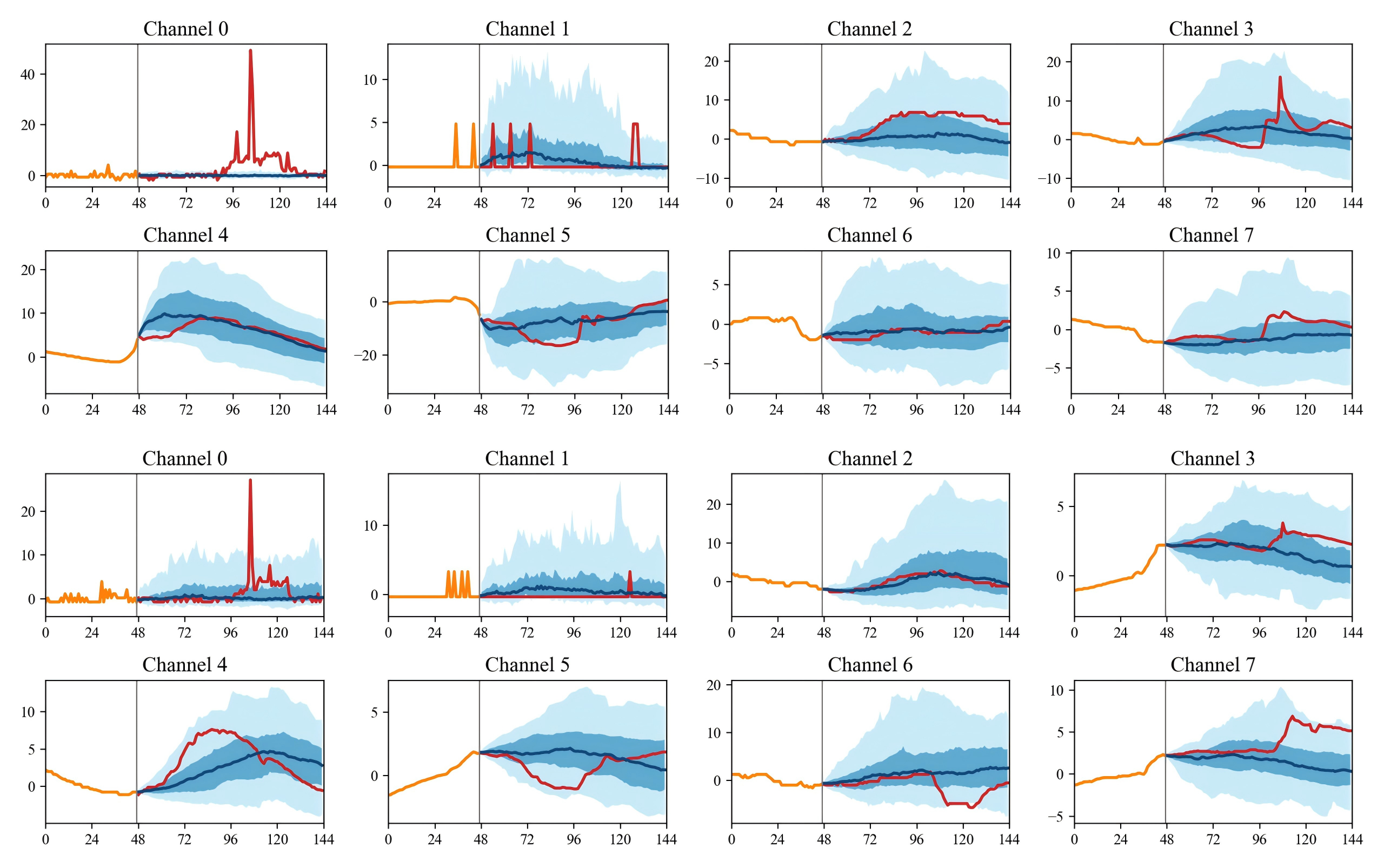}
\vspace{-10pt}
\caption{Appliance prediction intervals of first 8 channels.}
\vspace{-10pt}
\label{appliance_pi}
\end{figure}

\begin{figure}[ht]
\centering
\includegraphics[width=1.0\textwidth]{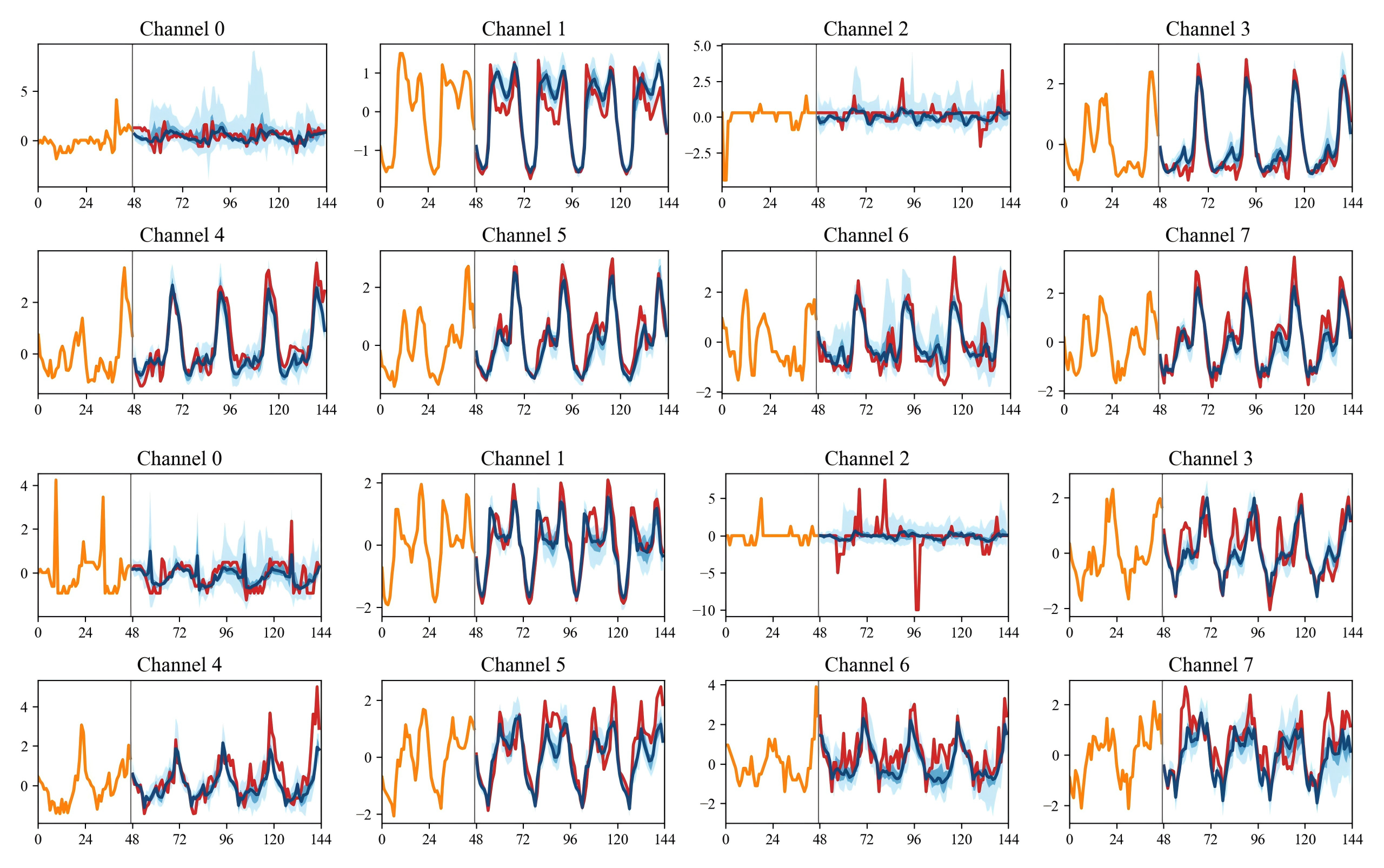}
\vspace{-10pt}
\caption{Electricity prediction intervals of first 8 channels.}
\vspace{-10pt}
\label{electricity_pi}
\end{figure}

\begin{figure}[ht]
\centering
\includegraphics[width=1.0\textwidth]{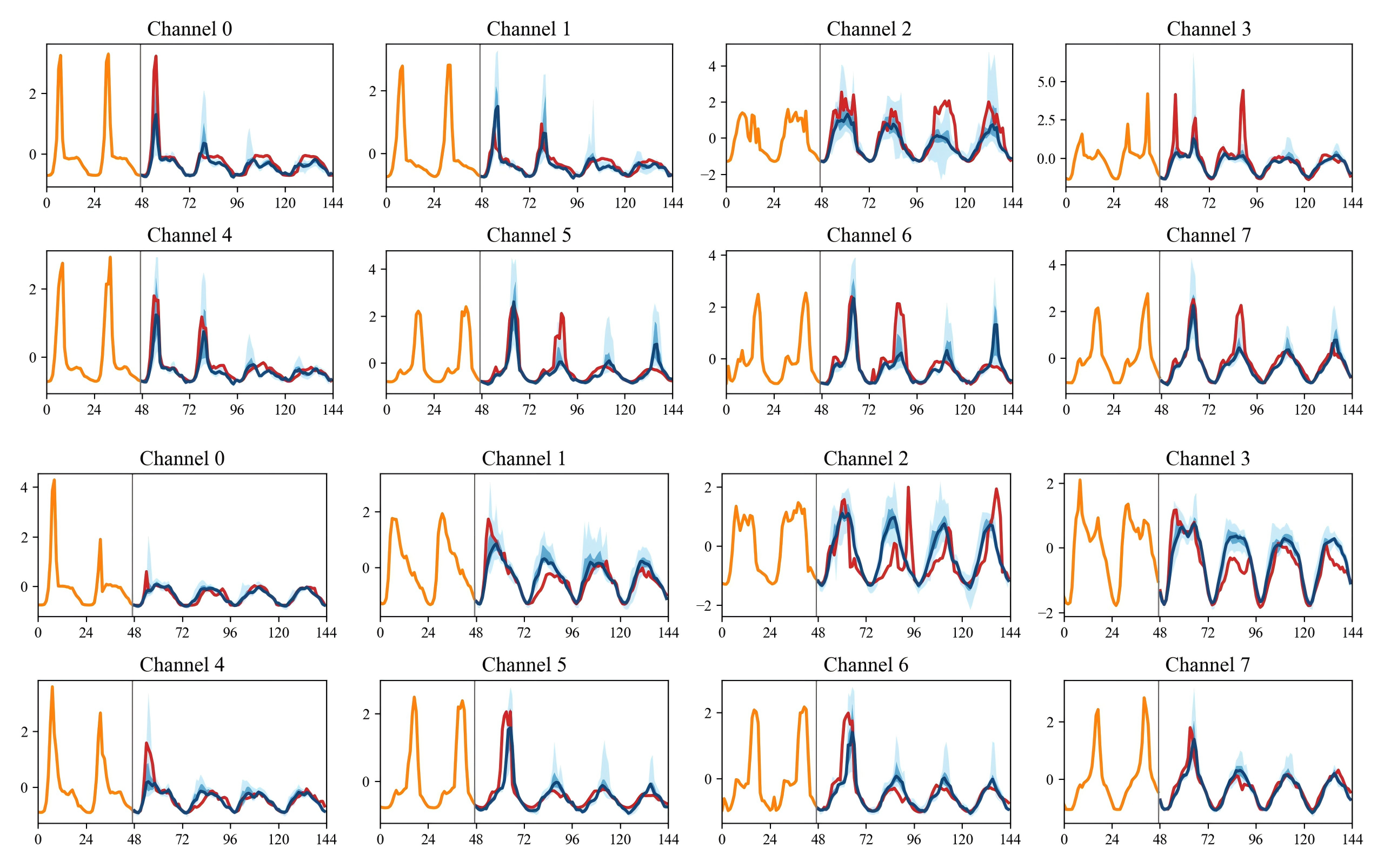}
\vspace{-10pt}
\caption{Traffic prediction intervals of first 8 channels.}
\vspace{-10pt}
\label{traffic_pi}
\end{figure}

%% file: proof.tex
Below, we shed light on how to derive the upper bound of diffusion-induced probabilistic forecasting error shown in Proposition 1. We utilize the KL-divergence between the real distribution $q^{te}(\mathbf{y}_{0}|\mathbf{x})$ of test time series and approximated predictive distribution $p_{\theta}^{te}(\mathbf{y}_{0}|\mathbf{x})$ by conditional diffusion models to represent the probabilistic forecasting error
\begin{equation}
\label{eq:8}
\mathcal{D}_{KL}\left[q^{te}(\mathbf{y}_{0}|\mathbf{x})||p^{te}_{\theta}(\mathbf{y}_{0}|\mathbf{x})\right] = \mathbb{E}_{q^{te}(\mathbf{y}_{0}|\mathbf{x})}\left[\log q^{te}(\mathbf{y}_{0}|\mathbf{x})\right] - \mathbb{E}_{q^{te}(\mathbf{y}_{0}|\mathbf{x})}\left[\log p^{te}_{\theta}(\mathbf{y}_{0}|\mathbf{x})\right]. 
\end{equation}

The first term in Eq. \ref{eq:8} is unrelated to conditional diffusion learning and thus can be prescribed as a constant $C_{1}$ based on the information quantity of real test data
\begin{equation}
\label{eq:9}
\mathbb{E}_{q^{te}(\mathbf{y}_{0}|\mathbf{x})}\left[\log q^{te}(\mathbf{y}_{0}|\mathbf{x})\right] = -\frac{1}{q^{te}(\mathbf{x})} \mathbb{E}_{q^{te}(\mathbf{y}_{0}, \mathbf{x})}\left[\log q^{te}(\mathbf{y}_{0}|\mathbf{x})\right] = -\frac{\mathit{H}(\mathbf{y}_{0}|\mathbf{x})}{q^{te}(\mathbf{x})} = \mathrm{C}_{1}.
\end{equation}

The second term Eq. \ref{eq:8} is the expected log-likelihood over $q^{te}(\mathbf{y}_{0}|\mathbf{x})$, which is identical to the learning objective of vanilla conditional diffusion models in \citep{ho2020denoising}. Akin to the step-wise denoising loss derivation in \citep{ho2020denoising}, we can obtain the upper bound of the error via Jensen's inequality and decompose it into $K+1$ items $\mathcal{V}_{0},...,\mathcal{V}_{K}$:
\begin{align}
\label{eq:10}
-\mathbb{E}_{q^{te}(\mathbf{y}_{0}|\mathbf{x})}\left[\log p^{te}_{\theta}(\mathbf{y}_{0}|\mathbf{x})\right] & = -\mathbb{E}_{q^{te}(\mathbf{y}_{0}|\mathbf{x})}\left[\log \int q^{te}(\mathbf{y}_{1:K}|\mathbf{y}_{0}) \frac{p^{te}_{\theta}(\mathbf{y}_{0:K}|\mathbf{x})}{q^{te}(\mathbf{y}_{1:K}|\mathbf{y}_{0})}\mathrm{d}\mathbf{y}_{1:K}\right] \nonumber \\
& \le -\mathbb{E}_{q^{te}(\mathbf{y}_{0}|\mathbf{x})}\left[\mathbb{E}_{q^{te}(\mathbf{y}_{1:K}|\mathbf{y}_{0})}\left[\log \frac{p^{te}_{\theta}(\mathbf{y}_{0:K}|\mathbf{x})}{q^{te}(\mathbf{y}_{1:K}|\mathbf{y}_{0})}\right]\right] \nonumber \\
& = \mathbb{E}_{q^{te}(\mathbf{y}_{0}|\mathbf{x})}\left[\mathcal{V}_{0}+\sum_{k=2}^{K}\mathcal{V}_{k-1}+\mathcal{V}_{K}\right],
\end{align}
where
\begin{align}
\label{eq:11}
\mathcal{V}_{K}=\mathcal{D}_{KL}\left[q^{te}(\mathbf{y}_{K}|\mathbf{y}_{0})||p^{te}_{\theta}(\mathbf{y}_{K}|\mathbf{x})\right]=0,
\end{align}
as $q^{te}(\mathbf{y}_{K}|\mathbf{y}_{0})$ and $p^{te}_{\theta}(\mathbf{y}_{K}|\mathbf{x})$ are both standard Gaussian. And since the reverse transitions at each diffusion step can be shaped in explicit Gaussian forms, we can write out
\begin{align}
\label{eq:12}
\mathcal{V}_{k-1} & = \mathbb{E}_{q^{te}(\mathbf{y}_{k}|\mathbf{y}_{0})}\left[\mathcal{D}_{KL}\left[q^{te}(\mathbf{y}_{k-1}|\mathbf{y}_{k}, \mathbf{y}_{0})||p^{te}_{\theta}(\mathbf{y}_{k-1}|\mathbf{y}_{k}, \mathbf{x})\right]\right] \nonumber \\
& = \mathbb{E}_{q^{te}(\mathbf{y}_{k}|\mathbf{y}_{0})}\left[\mathcal{D}_{KL}\left[\mathcal{N}(\mathbf{y}_{k-1}; \bm{\mu}_{k}(\mathbf{y}_{k},\mathbf{y}_{0}), \tilde{\beta}_{k}\mathbf{I})||\mathcal{N}(\mathbf{y}_{k-1}; \bm{\mu}_{\theta}(\mathbf{y}_{k},\mathbf{x}, k), \tilde{\beta}_{k}\mathbf{I})\right]\right] \nonumber \\
& = \mathbb{E}_{q^{te}(\mathbf{y}_{k}|\mathbf{y}_{0})}\left[\frac{1}{2\tilde{\beta}_{k}^{2}}\left[\left\|\bm{\mu}_{\theta}(\mathbf{y}_{k},\mathbf{x}, k)-\bm{\mu}_{k}(\mathbf{y}_{k},\mathbf{y}_{0})\right\|^{2}_{2}\right]\right] \nonumber \\
& = \mathbb{E}_{\mathbf{y}_{0},\bm{\epsilon}_{k}}\left[\frac{1}{2\tilde{\beta}_{k}^{2}}\left[\left\|\frac{1}{\sqrt{\alpha_{k}}}\left(\mathbf{y}_{k}-\frac{\beta_{k}}{\sqrt{1-\bar{\alpha}_{k}}} \bm{\epsilon}_{\theta}\left(\mathbf{y}_{k}, \mathbf{x}, k\right)\right)-\frac{1}{\sqrt{\alpha_{k}}}\left(\mathbf{y}_{k}-\frac{\beta_{k}}{\sqrt{1-\bar{\alpha}_{k}}} \bm{\epsilon}_{k}\right)\right\|^{2}_{2}\right]\right] \nonumber \\
& = \mathbb{E}_{\mathbf{y}_{0},\bm{\epsilon}_{k}}\left[\frac{\beta_{k}^{2}}{2\tilde{\beta}_{k}^{2}\alpha_{k}(1-\bar{\alpha}_{k})}\left[\left\|\bm{\epsilon}_{\theta}\left(\sqrt{\bar{\alpha}_{k}}\mathbf{y}_{0}+\sqrt{1-\bar{\alpha}_{k}}\bm{\epsilon}_{k}, \mathbf{x}, k\right)-\bm{\epsilon}_{k}\right\|^{2}_{2}\right]\right],
\end{align}
where $\tilde{\beta}_{k}=\frac{1-\tilde{\alpha}_{k-1}}{1-\tilde{\alpha}_{k}}{\beta}_{k}$
and $\mathcal{V}_{0}$ is actually a special case of Eq. \ref{eq:12} when $k=1$
\begin{align}
\label{eq:13}
\mathcal{V}_{0} & = -\mathbb{E}_{q(\mathbf{y}_{1}|\mathbf{y}_{0})}\left[ \log p_{\theta}(\mathbf{y}_{0}|\mathbf{y}_{1}, \mathbf{x})\right] \nonumber \\
& = \mathbb{E}_{q(\mathbf{y}_{1}|\mathbf{y}_{0})}\left[ \log (2\pi)^\frac{HD}{2}\tilde{\beta}_{1}+\frac{1}{2\tilde{\beta}_{1}^{2}}\left\| \mathbf{y}_{0}-\bm{\mu}_{\theta}(\mathbf{y}_{1},\mathbf{x}, k=1)\right\|^{2}_{2} \right] \nonumber \\
& = \mathbb{E}_{\mathbf{y}_{0},\bm{\epsilon}_{1}}\left[ \frac{\beta_{1}^{2}}{2\tilde{\beta}_{1}^{2}\alpha_{1}(1-\bar{\alpha}_{1})}\left[\left\|\bm{\epsilon}_{\theta}\left(\sqrt{\bar{\alpha}_{1}}\mathbf{y}_{0}+\sqrt{1-\bar{\alpha}_{1}}\bm{\epsilon}_{1}, \mathbf{x}, k=1\right)-\bm{\epsilon}_{1}\right\|^{2}_{2}\right] \right]+\mathrm{C}_{2}.
\end{align}

Overall, if we let $A_{k}=\frac{\beta_{k}^{2}}{2\tilde{\beta}_{k}^{2}\alpha_{k}(1-\bar{\alpha}_{k})}$, $C=C_{1}+C_{2}$, we can derive the ultimate upper bound of probabilistic forecasting error in a concise form as follows:
\begin{equation}
\centering
\mathcal{D}_{KL}\left[q^{te}(\mathbf{y}_{0}|\mathbf{x})||p^{te}_{\theta}(\mathbf{y}_{0}|\mathbf{x})\right] \le \mathbb{E}_{\mathbf{x},\mathbf{y}_{0},\bm{\epsilon}_{k},k}\left[A_{k}\left\|\bm{\epsilon}_{\theta}\left(\sqrt{\bar{\alpha}_{k}}\mathbf{y}_{0}+\sqrt{1-\bar{\alpha}_{k}}\bm{\epsilon}_{k}, \mathbf{x}, k\right)-\bm{\epsilon}_{k}\right\|^{2}_{2}\right]+C,
\label{eq:14}
\end{equation}
which finalizes the proof of Proposition 1. It shows that for unknown test time series, the diffusion-based generative forecasting performance is associated with the generalization capability of the trained conditional denoising network on total step-wise noise regression.

%% file: related_work.tex
\textbf{Channel-oriented multivariate forecasting.} How to properly manage various channel-centric temporal properties (i.e. single-channel dynamics and cross-channel correlations) has been attached greater importance in recent multivariate forecasting works \citep{chen2024similarity, han2024softs} for two reasons. One is that traditional transformer-based models \citep{zhou2021informer, wu2021autoformer, zhou2022fedformer, liu2022non} only focus on improving the expressivity and efficiency of long-range temporal dependency, which can not obviously discriminate roles of disparate channels and entice some unsatisfactory outcomes. Besides, channel-independent predictors \citep{nie2022time, zeng2023transformers, das2023long} utilize a shared network to uniformly treat all channels and display that the single-channel separate prediction can outperform multi-channel mixing settings. Whilst this channel-independent structure fail to handle those complex temporal modes where the auxiliary information from other channels could also be helpful. Latest progress \citep{liu2023itransformer, lu2023arm, chen2024similarity, han2024softs} reflect that both channel-independence and channel-fusion are crucial for versatile time series predictors. However, the significance of proper channel manipulation is rarely probed in multivariate diffusion forecasters, and the additional influence of channel noise imposed in different extents should also be considered. To tackle this barrier, we blend both channel-independence and channel-fusion modules in diffusion denoiser to boost its forecasting ability on multivariate cases.

\textbf{Time series diffusion models.} Due to the remarkable capacity to generate high-fidelity samples, diffusion models are actively exploited to grasp the stochastic dynamics and temporal correlations for a variety of time series tasks, including synthesis \citep{yuan2024diffusion, narasimhan2024time}, forecasting \citep{rasul2021autoregressive}, imputation \citep{tashiro2021csdi} and anomaly detection \citep{chen2023imdiffusion}. Common goals of these tasks are to derive a high-quality conditional temporal distribution aligned with diverse input contexts, such as statistical properties in constrained generation \citep{coletta2024constrained} and historical records. To this end, the key challenge lies in how to design a potent temporal conditioning mechanism to empower the conditional backward generation. An intuitive way is to integrate useful temporal properties such as trend-seasonality \citep{yuan2024diffusion}, continuity \citep{bilovs2023modeling} and multi-scale modes \citep{shen2023multi, fan2024mg} to empirically boost the utilization efficiency of conditioning data in the learnable denoising process. Another track is to develop gradient-based guidance schemes to satisfy given constraints via differentialable \citep{coletta2024constrained} or objective-oriented optimization \citep{kollovieh2024predict}. Even this plethora of time series diffusion models, there are still rooms to enhance them from the aspect of training manners and denoiser architectures. To bridge this gap for multivariate forecasting, we exclusively design a channel-aware denoiser network and recast the problem of estimating conditional predictive distribution in the paradigm of mutual information maximization, which can enhance the consistency between past conditioning and future predicted time series. On top of original conditional likelihood maximization via step-wise noise regression, we adapt temporal contrastive learning to further augment conditional diffusion training. In future work, we hope to extend such innovations to benefit other time series analysis tasks.

\textbf{Temporal contrastive learning.} Time series contrastive learning primarily aims to obtain self-supervised universal temporal representations which can enable an array of downstream tasks with few shots \citep{trirat2024universal, lee2023soft, franceschi2019unsupervised, wang2024contrast}. This line of research focus on developing efficient representation learning methods to pre-train temporal feature extractors in two vital senses, containing contrastive loss design and positive and negative sample pair construction. With respect to the deterministic time series prediction task, there also exist specialized decomposed contrastive pre-training approaches \citep{woo2021cost, wang2022learning} to investigate disentangled seasonal and trend representations, which can relieve the subsequent prediction on volatile temporal evolution. While in this work, we devise an end-to-end denoising-based contrastive learning to ameliorate conditional denoiser training rather than the common pre-training fashion on general temporal representation networks. We realize this contrastive refinement in an identical form of step-wise noise regression to vanilla diffusion. Moreover, we alter both temporal variations and point magnitudes in the time series augmentation stage, which can construct more useful negative samples for the contrastive denoiser improvement.

%% file: train_algo.tex
\begin{algorithm}[h]
 \caption{Step-wise contrastive conditional diffusion training procedure.}
\label{train_algo}
 \textbf{Input:} Lookback time series $\mathbf{x}\in \mathbb{R}^{L\times D}$; target time series $\mathbf{y}_{0}\in \mathbb{R}^{H\times D}$; lookback length $L$; prediction horizon $H$; variate number $D$; diffusion step number $K$; negative sample number $N$; contrastive loss weight $\lambda$; temperature coefficient $\tau$; \\
 \textbf{repeat}
 \begin{algorithmic}[1]
 \State Draw step $k \sim \mathbb{U}[1,..,K]$.
 \State Draw noise $\bm{\epsilon}\sim \mathcal{N}(\mathbf{0}, \mathbf{I})$ to calculate the naive diffusion loss $\mathcal{L}_{k}^{denoise}$ in Eq. \ref{eq:1}.
 \State Draw noise $\bm{\epsilon}{}'\sim \mathcal{N}(\mathbf{0}, \mathbf{I})$ to calculate the denoising-based contrastive loss $\mathcal{L}_{k}^{contrast}$ in Eq. \ref{eq:4}.
 \State Obtain a set of negative time series $\left \{ \mathbf{y}_{0}^{n} \right \}_{n=1}^{N}$ using the hybrid augmentation in Appendix \ref{neg_aug}.
 \State Compute the contrastive conditional diffusion loss $\mathcal{L}^{CCDM}_{k}=\mathcal{L}_{k}^{denoise}+\lambda \mathcal{L}_{k}^{contrast}$ in Eq. \ref{eq:6}.
 \State Optimize the conditional denoising network $\bm{\epsilon}_{\theta}(\cdot)$ using the gradient $\nabla_{\theta} \mathcal{L}^{CCDM}_{k}$.
\end{algorithmic}
\textbf{until} converged
\end{algorithm}

%% file: dataset_table.tex
\begin{table}[!h]
\centering
\caption{Detailed dataset description. Size indicates the split lengths of individual points for training, validation and testing division respectively.}
\label{dataset_table}
\begin{tabular}{c|c|c|c|c}
\toprule
Dataset     & Variate number $D$ & Sampling frequency & Split size         & Field                                                                                \\ \midrule
ETTh1       & 7              & Hourly    & (8640, 2880, 2880)   & Energy                                                \\
Exchange    & 8              & Daily     & (5311, 758, 1517)    & Finance \\
Weather     & 21             & 10min     & (34560, 5760, 11520) & Weather                                                   \\
Appliance   & 28             & 10min     & (13814, 1973, 3947)  & Energy                  \\
Electricity & 321            & Hourly    & (17280, 2880, 5760)  & Energy               \\
Traffic     & 862            & Hourly    & (11520, 2880, 2880)  & Traffic              \\ 
\bottomrule
\end{tabular}
\end{table}

%% file: config_table.tex
\begin{table}[!h]
\centering
\caption{Diffusion forecaster configurations on different forecasting setups.}
\label{config_table}
\resizebox{1.0\textwidth}{!}{
\begin{tabular}{cc|ccc|ccc}
\toprule
\multicolumn{2}{c|}{Forecasting setup} & \multicolumn{3}{c|}{DiT blocks} & \multicolumn{3}{c}{\begin{tabular}[c]{@{}c@{}}Noise schedule\\ (quadratic)\end{tabular}} \\ \midrule
Lookback length $L$  & Prediction horizon $H$  & Depth $n_{att}$   & Heads   & Hidden dim $e_{hid}$  & $\beta_{1}$                        & $\beta_{K}$                       & Steps $K$                      \\ \midrule
48               & 96                  & 2       & 8       & 128         & 0.0001                       & 0.5                         & 50                          \\
96               & 168                 & 2       & 8       & 256         & 0.0001                       & 0.2                         & 100                         \\
192              & 336                 & 2       & 8       & 512         & 0.0001                       & 0.1                         & 200                         \\
336              & 720                 & 2       & 8       & 728         & 0.0001                       & 0.1                         & 200                         \\ 
\bottomrule
\end{tabular}
}
\end{table}

%% file: implement_table.tex
\begin{table}[!h]
\centering
\caption{Contrastive training configurations corresponding to forecasting results in Table \ref{overall_results}.}
\label{implement_table}
\resizebox{1.0\textwidth}{!}{
\begin{tabular}{cc|ccccc}
\toprule
\multicolumn{2}{c|}{Setup}         & Contrastive weight $\lambda$ & Negative number $N$ & Batch size & Initial rate & Training mode \\ \midrule
\multirow{4}{*}{ETTh1}       & 96  & 0.001              & 64*2            & 32         & 0.001        & End-to-end    \\
                             & 168 & 0.001              & 64*2            & 32         & 0.001        & End-to-end    \\
                             & 336 & 0.001              & 64*2            & 32         & 0.0002       & End-to-end    \\
                             & 720 & 0.001              & 64*2            & 32         & 0.0002       & End-to-end    \\ \midrule
\multirow{4}{*}{Exchange}    & 96  & 0.001              & 64*2            & 32         & 0.001        & End-to-end    \\
                             & 168 & 0.001              & 64*2            & 32         & 0.001        & End-to-end    \\
                             & 336 & 0.001              & 64*2            & 32         & 0.0002       & End-to-end    \\
                             & 720 & 0.001              & 64*2            & 32         & 0.0002       & End-to-end    \\ \midrule
\multirow{4}{*}{Weather}     & 96  & 0.001              & 64*2            & 32         & 0.0001       & Two-stage     \\
                             & 168 & 0.001              & 64*2            & 32         & 0.0001       & Two-stage     \\
                             & 336 & 0.0001             & 64*2            & 32         & 0.00002      & Two-stage     \\
                             & 720 & 0.001              & 64*2            & 32         & 0.00002      & Two-stage     \\ \midrule
\multirow{4}{*}{Appliance}   & 96  & 0.001              & 64*2            & 32         & 0.001        & End-to-end    \\
                             & 168 & 0.001              & 64*2            & 32         & 0.001        & End-to-end    \\
                             & 336 & 0.0001             & 64*2            & 32         & 0.0002       & End-to-end    \\
                             & 720 & 0.00005            & 64*2            & 32         & 0.0002       & End-to-end    \\ \midrule
\multirow{4}{*}{Electricity} & 96  & 0.0001             & 32*2            & 16         & 0.0001       & Two-stage     \\
                             & 168 & 0.00005            & 32*2            & 12         & 0.0001       & Two-stage     \\
                             & 336 & 0.00005            & 32*2            & 8          & 0.00002      & Two-stage     \\
                             & 720 & 0.00005            & 20*2            & 8          & 0.00002      & Two-stage     \\ \midrule
\multirow{4}{*}{Traffic}     & 96  & 0.0005             & 28*2            & 4          & 0.0001       & Two-stage     \\
                             & 168 & 0.0001             & 22*2            & 4          & 0.0001       & Two-stage     \\
                             & 336 & 0.00005            & 16*2            & 4          & 0.00002      & Two-stage     \\
                             & 720 & 0.00005            & 12*2            & 4          & 0.00002      & Two-stage     \\ 
\bottomrule
\end{tabular}
}
\end{table}

%% file: time_table.tex
\begin{table}[t]
\centering
\caption{Time cost comparison of diffusion forecasters on different sizes of prediction tasks. Both training time [s] of one epoch and inference time [ms] of one step are provided.}
\label{time_table}
\resizebox{1.0\textwidth}{!}{
\begin{tabular}{cc|cc|cc|cc|cc}
\toprule
\multicolumn{2}{c|}{\multirow{2}{*}{Size}} & \multicolumn{2}{c|}{CCDM}      & \multicolumn{2}{c|}{TimeDiff}  & \multicolumn{2}{c|}{CSDI}      & \multicolumn{2}{c}{TimeGrad}   \\ \cmidrule{3-10} 
\multicolumn{2}{c|}{}                       & Train {[}s{]} & Infer {[}ms{]} & Train {[}s{]} & Infer {[}ms{]} & Train {[}s{]} & Infer {[}ms{]} & Train {[}s{]} & Infer {[}ms{]} \\ \midrule
\multirow{4}{*}{D=8}          & H=96        & 18.67         & 3.63           & 14.11         & 3.00           & 4.78          & 3.63           & 2.22          & 349.42         \\
                              & H=168       & 28.11         & 4.37           & 18.56         & 3.00           & 6.33          & 3.58           & 3.89          & 603.05         \\
                              & H=336       & 80.78         & 4.71           & 24.00         & 2.98           & 10.67         & 3.72           & 5.32          & 1163.55        \\
                              & H=720       & 166.44        & 4.97           & 26.22         & 3.05           & 18.78         & 3.58           & 9.33          & 2571.37        \\ \midrule
\multirow{4}{*}{D=28}         & H=96        & 66.67         & 3.76           & 25.44         & 4.00           & 34.89         & 3.76           & 7.67          & 374.23         \\
                              & H=168       & 203.11        & 4.38           & 37.11         & 3.94           & 50.78         & 3.68           & 12.22         & 605.80         \\
                              & H=336       & 441.74        & 4.71           & 33.22         & 4.29           & 97.22         & 3.66           & 21.67         & 1170.64        \\
                              & H=720       & 903.00        & 4.75           & 34.67         & 4.50           & 181.56        & 6.45           & 50.22         & 2551.13        \\ \midrule
\multirow{4}{*}{D=321}        & H=96        & 573.22        & 4.59           & 657.67        & 17.92          & 84.78         & 9.08           & 48.56         & 357.51         \\
                              & H=168       & 1131.89       & 4.70           & 859.44        & 19.48          & 145.89        & 17.20          & 86.22         & 630.14         \\
                              & H=336       & 3173.89       & 4.83           & 1190.33       & 20.61          & 376.11        & 47.77          & 171.44        & 1188.07        \\
                              & H=720       & 4039.56       & 5.09           & 1269.56       & 22.71          & 546.67        & 70.13          & 330.67        & 2672.31        \\ \midrule
\multirow{4}{*}{D=862}        & H=96        & 1466.14       & 4.54           & 185.78        & 46.67          & 104.22        & 25.12          & 80.56         & 369.04         \\
                              & H=168       & 1884.77       & 4.52           & 193.89        & 47.89          & 118.33        & 47.71          & 146.67        & 620.23         \\
                              & H=336       & 3202.85       & 5.17           & 284.89        & 49.09          & 228.44        & 96.06          & 289.11        & 1186.83        \\
                              & H=720       & 4678.78       & 7.86           & 463.56        & 55.01          & 417.33        & 193.62         & 545.67        & 2591.93        \\ 
\bottomrule
\end{tabular}
}
\end{table}

%% file: ablation_table.tex
\begin{table}[t]
\centering
\caption{Complete forecasting results by masking denoising-based temporal contrastive refinement or channel-mixing DiT blocks.}
\label{ablation_table}
\resizebox{1.0\textwidth}{!}{
\begin{tabular}{cc|cccc|cccc}
\toprule
\multicolumn{2}{c|}{Methods}                            & \multicolumn{4}{c|}{w/o contrastive refinement}                             & \multicolumn{4}{c}{w/o channel-wise DiT}                                      \\ \midrule
\multicolumn{2}{c|}{Metrics}                            & MSE    & \multicolumn{1}{c|}{Degradation} & CRPS   & Degradation & MSE    & \multicolumn{1}{c|}{Degradation} & CRPS   & Degradation \\ \midrule
\multicolumn{1}{c|}{\multirow{5}{*}{\rotatebox[origin=c]{90}{ETTh1}}}       & 96  & 0.4447 & \multicolumn{1}{c|}{15.33\%}     & 0.3199 & 8.99\%      & 0.3903 & \multicolumn{1}{c|}{1.22\%}      & 0.2963 & 0.95\%      \\
\multicolumn{1}{c|}{}                             & 168 & 0.5223 & \multicolumn{1}{c|}{22.40\%}     & 0.3402 & 8.27\%      & 0.5800 & \multicolumn{1}{c|}{35.93\%}     & 0.6674 & 112.41\%    \\
\multicolumn{1}{c|}{}                             & 336 & 0.6416 & \multicolumn{1}{c|}{20.78\%}     & 0.3917 & 13.47\%     & 0.5381 & \multicolumn{1}{c|}{1.30\%}      & 0.4699 & 36.12\%     \\
\multicolumn{1}{c|}{}                             & 720 & 0.5944 & \multicolumn{1}{c|}{5.35\%}      & 0.5038 & 3.45\%      & 0.8740 & \multicolumn{1}{c|}{54.91\%}     & 0.8928 & 83.33\%     \\ \cmidrule{2-10} 
\multicolumn{1}{c|}{}                             & Avg & 0.5508 & \multicolumn{1}{c|}{15.97\%}     & 0.3889 & 8.55\%      & 0.5956 & \multicolumn{1}{c|}{23.34\%}     & 0.5816 & 58.20\%     \\ \midrule
\multicolumn{1}{c|}{\multirow{5}{*}{\rotatebox[origin=c]{90}{Exchange}}}    & 96  & 0.1057 & \multicolumn{1}{c|}{16.80\%}     & 0.1677 & 8.54\%      & 0.0959 & \multicolumn{1}{c|}{5.97\%}      & 0.1598 & 3.43\%      \\
\multicolumn{1}{c|}{}                             & 168 & 0.1986 & \multicolumn{1}{c|}{21.25\%}     & 0.2338 & 8.29\%      & 0.2200 & \multicolumn{1}{c|}{34.31\%}     & 0.2777 & 28.62\%     \\
\multicolumn{1}{c|}{}                             & 336 & 0.4532 & \multicolumn{1}{c|}{2.84\%}      & 0.3557 & 1.14\%      & 0.4735 & \multicolumn{1}{c|}{7.44\%}      & 0.3870 & 10.04\%     \\
\multicolumn{1}{c|}{}                             & 720 & 1.2290 & \multicolumn{1}{c|}{5.18\%}      & 0.6038 & 2.97\%      & 1.1802 & \multicolumn{1}{c|}{1.00\%}      & 0.5975 & 1.89\%      \\ \cmidrule{2-10} 
\multicolumn{1}{c|}{}                             & Avg & 0.4966 & \multicolumn{1}{c|}{11.52\%}      & 0.3403 & 5.24\%      & 0.4924 & \multicolumn{1}{c|}{12.18\%}      & 0.3555 & 11.00\%      \\ \midrule
\multicolumn{1}{c|}{\multirow{5}{*}{\rotatebox[origin=c]{90}{Weather}}}     & 96  & 0.2825 & \multicolumn{1}{c|}{5.84\%}      & 0.1936 & 1.68\%      & 0.2919 & \multicolumn{1}{c|}{9.37\%}      & 0.2012 & 5.67\%      \\
\multicolumn{1}{c|}{}                             & 168 & 0.3349 & \multicolumn{1}{c|}{34.55\%}     & 0.2167 & 8.03\%      & 0.4199 & \multicolumn{1}{c|}{68.70\%}     & 0.2981 & 48.60\%     \\
\multicolumn{1}{c|}{}                             & 336 & 0.2932 & \multicolumn{1}{c|}{2.16\%}      & 0.2313 & 2.44\%      & 0.3825 & \multicolumn{1}{c|}{33.28\%}     & 0.2873 & 27.24\%     \\
\multicolumn{1}{c|}{}                             & 720 & 0.6158 & \multicolumn{1}{c|}{9.44\%}      & 0.4365 & 6.91\%      & 0.8428 & \multicolumn{1}{c|}{49.78\%}     & 0.5478 & 34.17\%     \\ \cmidrule{2-10} 
\multicolumn{1}{c|}{}                             & Avg & 0.3816 & \multicolumn{1}{c|}{13.00\%}     & 0.2695 & 4.77\%      & 0.4843 & \multicolumn{1}{c|}{40.28\%}     & 0.3336 & 28.92\%     \\ \midrule
\multicolumn{1}{c|}{\multirow{5}{*}{\rotatebox[origin=c]{90}{Appliance}}}   & 96  & 0.7097 & \multicolumn{1}{c|}{1.56\%}      & 0.4291 & 3.70\%      & 0.7473 & \multicolumn{1}{c|}{6.94\%}      & 0.4546 & 9.86\%      \\
\multicolumn{1}{c|}{}                             & 168 & 0.7313 & \multicolumn{1}{c|}{16.71\%}     & 0.4374 & 8.81\%      & 0.7853 & \multicolumn{1}{c|}{25.33\%}     & 0.6070 & 51.00\%     \\
\multicolumn{1}{c|}{}                             & 336 & 0.9254 & \multicolumn{1}{c|}{1.48\%}      & 0.5083 & 0.93\%      & 1.0660 & \multicolumn{1}{c|}{16.90\%}     & 0.6971 & 38.42\%     \\
\multicolumn{1}{c|}{}                             & 720 & 1.7215 & \multicolumn{1}{c|}{8.97\%}      & 0.9525 & 10.50\%     & 1.8744 & \multicolumn{1}{c|}{18.65\%}     & 1.1336 & 31.51\%     \\ \cmidrule{2-10} 
\multicolumn{1}{c|}{}                             & Avg & 1.0220 & \multicolumn{1}{c|}{7.18\%}      & 0.5818 & 5.99\%      & 1.1183 & \multicolumn{1}{c|}{16.96\%}     & 0.7231 & 32.70\%     \\ \midrule
\multicolumn{1}{c|}{\multirow{5}{*}{\rotatebox[origin=c]{90}{Electricity}}} & 96  & 0.2142 & \multicolumn{1}{c|}{1.90\%}      & 0.2266 & 3.85\%      & 0.2296 & \multicolumn{1}{c|}{9.23\%}      & 0.2198 & 0.73\%      \\
\multicolumn{1}{c|}{}                             & 168 & 0.1689 & \multicolumn{1}{c|}{0.66\%}      & 0.2033 & 0.94\%      & 0.1779 & \multicolumn{1}{c|}{6.02\%}      & 0.2041 & 1.34\%      \\
\multicolumn{1}{c|}{}                             & 336 & 0.1714 & \multicolumn{1}{c|}{1.84\%}      & 0.2035 & 1.04\%      & 0.1744 & \multicolumn{1}{c|}{3.62\%}      & 0.2048 & 1.69\%      \\
\multicolumn{1}{c|}{}                             & 720 & 0.2002 & \multicolumn{1}{c|}{0.40\%}      & 0.2242 & 0.45\%      & 0.2073 & \multicolumn{1}{c|}{3.96\%}      & 0.2260 & 1.25\%      \\ \cmidrule{2-10} 
\multicolumn{1}{c|}{}                             & Avg & 0.1887 & \multicolumn{1}{c|}{1.20\%}      & 0.2144 & 1.57\%      & 0.1973 & \multicolumn{1}{c|}{5.71\%}      & 0.2137 & 1.25\%      \\ \midrule
\multicolumn{1}{c|}{\multirow{5}{*}{\rotatebox[origin=c]{90}{Traffic}}}     & 96  & 1.0345 & \multicolumn{1}{c|}{0.61\%}      & 0.4226 & 3.25\%      & 1.2831 & \multicolumn{1}{c|}{24.79\%}     & 0.4741 & 15.83\%     \\
\multicolumn{1}{c|}{}                             & 168 & 0.6936 & \multicolumn{1}{c|}{0.80\%}      & 0.3113 & 1.17\%      & 0.7682 & \multicolumn{1}{c|}{11.64\%}     & 0.3869 & 25.74\%     \\
\multicolumn{1}{c|}{}                             & 336 & 0.6913 & \multicolumn{1}{c|}{0.73\%}      & 0.3572 & 6.37\%      & 0.8472 & \multicolumn{1}{c|}{23.44\%}     & 0.4329 & 28.92\%     \\
\multicolumn{1}{c|}{}                             & 720 & 0.9561 & \multicolumn{1}{c|}{2.18\%}      & 0.5610 & 24.14\%     & 1.1351 & \multicolumn{1}{c|}{21.31\%}     & 0.5762 & 27.51\%     \\ \cmidrule{2-10} 
\multicolumn{1}{c|}{}                             & Avg & 0.8439 & \multicolumn{1}{c|}{1.08\%}      & 0.4130 & 8.73\%      & 1.0084 & \multicolumn{1}{c|}{20.30\%}     & 0.4675 & 24.50\%     \\ \bottomrule
\end{tabular}
}
\end{table}

%% file: temp_table.tex
\begin{table}[!t]
\centering
\caption{Forecasting results by different temperature coefficients.}
\label{temp_table}
\resizebox{0.8\textwidth}{!}{
\begin{tabular}{c|cc|cc|cc}
\toprule
\multirow{2}{*}{Temperature $\tau$} & \multicolumn{2}{c|}{ETTh1} & \multicolumn{2}{c|}{Exchange} & \multicolumn{2}{c}{Weather} \\ \cmidrule{2-7} 
                             & MSE          & CRPS        & MSE           & CRPS          & MSE          & CRPS         \\ \midrule
0.05                         & 0.4391       & 0.3124      & 0.1728        & 0.2216        & 0.2515       & 0.2015       \\
0.1                          & 0.4267       & 0.3142      & 0.1638        & 0.2159        & 0.2489       & 0.2006       \\
0.5                          & 0.4730       & 0.3252      & 0.1583        & 0.2152        & 0.2493       & 0.2022       \\
1.0                          & 0.5082       & 0.3389      & 0.2031        & 0.2449        & 0.2505       & 0.2025       \\ 
\bottomrule
\end{tabular}
}
\end{table}

%% file: main.bbl
\begin{thebibliography}{63}
\providecommand{\natexlab}[1]{#1}
\providecommand{\url}[1]{\texttt{#1}}
\expandafter\ifx\csname urlstyle\endcsname\relax
  \providecommand{\doi}[1]{doi: #1}\else
  \providecommand{\doi}{doi: \begingroup \urlstyle{rm}\Url}\fi

\bibitem[Alcaraz \& Strodthoff(2022)Alcaraz and Strodthoff]{alcaraz2022diffusion}
Juan~Lopez Alcaraz and Nils Strodthoff.
\newblock Diffusion-based time series imputation and forecasting with structured state space models.
\newblock \emph{Transactions on Machine Learning Research}, 2022.

\bibitem[Bilo{\v{s}} et~al.(2023)Bilo{\v{s}}, Rasul, Schneider, Nevmyvaka, and G{\"u}nnemann]{bilovs2023modeling}
Marin Bilo{\v{s}}, Kashif Rasul, Anderson Schneider, Yuriy Nevmyvaka, and Stephan G{\"u}nnemann.
\newblock Modeling temporal data as continuous functions with stochastic process diffusion.
\newblock In \emph{International Conference on Machine Learning}, pp.\  2452--2470. PMLR, 2023.

\bibitem[Cao et~al.(2024)Cao, Ye, and Liu]{cao2024timedit}
Defu Cao, Wen Ye, and Yan Liu.
\newblock Timedit: General-purpose diffusion transformers for time series foundation model.
\newblock In \emph{ICML 2024 Workshop on Foundation Models in the Wild}, 2024.

\bibitem[Chen et~al.(2024{\natexlab{a}})Chen, Lenssen, Feng, Hu, Fey, Tassiulas, Leskovec, and Ying]{chen2024similarity}
Jialin Chen, Jan~Eric Lenssen, Aosong Feng, Weihua Hu, Matthias Fey, Leandros Tassiulas, Jure Leskovec, and Rex Ying.
\newblock From similarity to superiority: Channel clustering for time series forecasting.
\newblock \emph{arXiv preprint arXiv:2404.01340}, 2024{\natexlab{a}}.

\bibitem[Chen et~al.(2024{\natexlab{b}})Chen, Mei, Fan, and Wang]{chen2024overview}
Minshuo Chen, Song Mei, Jianqing Fan, and Mengdi Wang.
\newblock An overview of diffusion models: Applications, guided generation, statistical rates and optimization.
\newblock \emph{arXiv preprint arXiv:2404.07771}, 2024{\natexlab{b}}.

\bibitem[Chen et~al.(2020)Chen, Kornblith, Norouzi, and Hinton]{chen2020simple}
Ting Chen, Simon Kornblith, Mohammad Norouzi, and Geoffrey Hinton.
\newblock A simple framework for contrastive learning of visual representations.
\newblock In \emph{International conference on machine learning}, pp.\  1597--1607. PMLR, 2020.

\bibitem[Chen et~al.(2023)Chen, Zhang, Ma, Liu, Ding, Li, He, Rajmohan, Lin, and Zhang]{chen2023imdiffusion}
Yuhang Chen, Chaoyun Zhang, Minghua Ma, Yudong Liu, Ruomeng Ding, Bowen Li, Shilin He, Saravan Rajmohan, Qingwei Lin, and Dongmei Zhang.
\newblock Imdiffusion: Imputed diffusion models for multivariate time series anomaly detection.
\newblock \emph{Proceedings of the VLDB Endowment}, 17\penalty0 (3):\penalty0 359--372, 2023.

\bibitem[Coletta et~al.(2024)Coletta, Gopalakrishnan, Borrajo, and Vyetrenko]{coletta2024constrained}
Andrea Coletta, Sriram Gopalakrishnan, Daniel Borrajo, and Svitlana Vyetrenko.
\newblock On the constrained time-series generation problem.
\newblock \emph{Advances in Neural Information Processing Systems}, 36, 2024.

\bibitem[Crabb{\'e} et~al.(2024)Crabb{\'e}, Huynh, Stanczuk, and van~der Schaar]{crabbetime}
Jonathan Crabb{\'e}, Nicolas Huynh, Jan~Pawel Stanczuk, and Mihaela van~der Schaar.
\newblock Time series diffusion in the frequency domain.
\newblock In \emph{Forty-first International Conference on Machine Learning}, 2024.

\bibitem[Das et~al.(2023{\natexlab{a}})Das, Kong, Leach, Mathur, Sen, and Yu]{das2023long}
Abhimanyu Das, Weihao Kong, Andrew Leach, Shaan~K Mathur, Rajat Sen, and Rose Yu.
\newblock Long-term forecasting with tide: Time-series dense encoder.
\newblock \emph{Transactions on Machine Learning Research}, 2023{\natexlab{a}}.

\bibitem[Das et~al.(2023{\natexlab{b}})Das, Kong, Sen, and Zhou]{das2023decoder}
Abhimanyu Das, Weihao Kong, Rajat Sen, and Yichen Zhou.
\newblock A decoder-only foundation model for time-series forecasting.
\newblock \emph{arXiv preprint arXiv:2310.10688}, 2023{\natexlab{b}}.

\bibitem[Deng et~al.(2024)Deng, Song, Tsang, and Xiong]{deng2024bigger}
Jinliang Deng, Xuan Song, Ivor~W Tsang, and Hui Xiong.
\newblock The bigger the better? rethinking the effective model scale in long-term time series forecasting.
\newblock \emph{arXiv preprint arXiv:2401.11929}, 2024.

\bibitem[Dumas et~al.(2022)Dumas, Wehenkel, Lanaspeze, Corn{\'e}lusse, and Sutera]{dumas2022deep}
Jonathan Dumas, Antoine Wehenkel, Damien Lanaspeze, Bertrand Corn{\'e}lusse, and Antonio Sutera.
\newblock A deep generative model for probabilistic energy forecasting in power systems: normalizing flows.
\newblock \emph{Applied Energy}, 305:\penalty0 117871, 2022.

\bibitem[Esser et~al.(2024)Esser, Kulal, Blattmann, Entezari, M{\"u}ller, Saini, Levi, Lorenz, Sauer, Boesel, et~al.]{esser2024scaling}
Patrick Esser, Sumith Kulal, Andreas Blattmann, Rahim Entezari, Jonas M{\"u}ller, Harry Saini, Yam Levi, Dominik Lorenz, Axel Sauer, Frederic Boesel, et~al.
\newblock Scaling rectified flow transformers for high-resolution image synthesis.
\newblock \emph{arXiv preprint arXiv:2403.03206}, 2024.

\bibitem[Fan et~al.(2024)Fan, Wu, Xu, Huang, Liu, and Bian]{fan2024mg}
Xinyao Fan, Yueying Wu, Chang Xu, Yuhao Huang, Weiqing Liu, and Jiang Bian.
\newblock Mg-tsd: Multi-granularity time series diffusion models with guided learning process.
\newblock \emph{arXiv preprint arXiv:2403.05751}, 2024.

\bibitem[Feng et~al.(2024)Feng, Miao, Zhang, and Zhao]{feng2024latent}
Shibo Feng, Chunyan Miao, Zhong Zhang, and Peilin Zhao.
\newblock Latent diffusion transformer for probabilistic time series forecasting.
\newblock In \emph{Proceedings of the AAAI Conference on Artificial Intelligence}, volume~38, pp.\  11979--11987, 2024.

\bibitem[Franceschi et~al.(2019)Franceschi, Dieuleveut, and Jaggi]{franceschi2019unsupervised}
Jean-Yves Franceschi, Aymeric Dieuleveut, and Martin Jaggi.
\newblock Unsupervised scalable representation learning for multivariate time series.
\newblock \emph{Advances in neural information processing systems}, 32, 2019.

\bibitem[Gao et~al.(2024)Gao, Chen, Wang, Wang, Wang, Gao, and Ding]{gao2024diffsformer}
Yuan Gao, Haokun Chen, Xiang Wang, Zhicai Wang, Xue Wang, Jinyang Gao, and Bolin Ding.
\newblock Diffsformer: A diffusion transformer on stock factor augmentation.
\newblock \emph{arXiv preprint arXiv:2402.06656}, 2024.

\bibitem[Han et~al.(2024)Han, Chen, Ye, and Zhan]{han2024softs}
Lu~Han, Xu-Yang Chen, Han-Jia Ye, and De-Chuan Zhan.
\newblock Softs: Efficient multivariate time series forecasting with series-core fusion.
\newblock \emph{arXiv preprint arXiv:2404.14197}, 2024.

\bibitem[Ho et~al.(2020)Ho, Jain, and Abbeel]{ho2020denoising}
Jonathan Ho, Ajay Jain, and Pieter Abbeel.
\newblock Denoising diffusion probabilistic models.
\newblock \emph{Advances in neural information processing systems}, 33:\penalty0 6840--6851, 2020.

\bibitem[Huang et~al.(2023)Huang, Wu, and Boulet]{huang2023metaprobformer}
Xingshuai Huang, Di~Wu, and Benoit Boulet.
\newblock Metaprobformer for charging load probabilistic forecasting of electric vehicle charging stations.
\newblock \emph{IEEE Transactions on Intelligent Transportation Systems}, 2023.

\bibitem[Ilbert et~al.(2024)Ilbert, Odonnat, Feofanov, Virmaux, Paolo, Palpanas, and Redko]{ilbert2024unlocking}
Romain Ilbert, Ambroise Odonnat, Vasilii Feofanov, Aladin Virmaux, Giuseppe Paolo, Themis Palpanas, and Ievgen Redko.
\newblock Unlocking the potential of transformers in time series forecasting with sharpness-aware minimization and channel-wise attention.
\newblock \emph{arXiv preprint arXiv:2402.10198}, 2024.

\bibitem[Jiang et~al.(2023)Jiang, Cornman, Park, Sapp, Zhou, Anguelov, et~al.]{jiang2023motiondiffuser}
Chiyu Jiang, Andre Cornman, Cheolho Park, Benjamin Sapp, Yin Zhou, Dragomir Anguelov, et~al.
\newblock Motiondiffuser: Controllable multi-agent motion prediction using diffusion.
\newblock In \emph{Proceedings of the IEEE/CVF Conference on Computer Vision and Pattern Recognition}, pp.\  9644--9653, 2023.

\bibitem[Kollovieh et~al.(2024)Kollovieh, Ansari, Bohlke-Schneider, Zschiegner, Wang, and Wang]{kollovieh2024predict}
Marcel Kollovieh, Abdul~Fatir Ansari, Michael Bohlke-Schneider, Jasper Zschiegner, Hao Wang, and Yuyang~Bernie Wang.
\newblock Predict, refine, synthesize: Self-guiding diffusion models for probabilistic time series forecasting.
\newblock \emph{Advances in Neural Information Processing Systems}, 36, 2024.

\bibitem[Lee et~al.(2023)Lee, Park, and Lee]{lee2023soft}
Seunghan Lee, Taeyoung Park, and Kibok Lee.
\newblock Soft contrastive learning for time series.
\newblock In \emph{The Twelfth International Conference on Learning Representations}, 2023.

\bibitem[Li et~al.(2024)Li, Carver, Lopez-Gomez, Sha, and Anderson]{li2024generative}
Lizao Li, Robert Carver, Ignacio Lopez-Gomez, Fei Sha, and John Anderson.
\newblock Generative emulation of weather forecast ensembles with diffusion models.
\newblock \emph{Science Advances}, 10\penalty0 (13):\penalty0 eadk4489, 2024.

\bibitem[Li et~al.(2022)Li, Lu, Wang, and Dou]{li2022generative}
Yan Li, Xinjiang Lu, Yaqing Wang, and Dejing Dou.
\newblock Generative time series forecasting with diffusion, denoise, and disentanglement.
\newblock \emph{Advances in Neural Information Processing Systems}, 35:\penalty0 23009--23022, 2022.

\bibitem[Li et~al.(2023)Li, Chen, Hu, Chen, Zhou, et~al.]{li2023transformer}
Yuxin Li, Wenchao Chen, Xinyue Hu, Bo~Chen, Mingyuan Zhou, et~al.
\newblock Transformer-modulated diffusion models for probabilistic multivariate time series forecasting.
\newblock In \emph{The Twelfth International Conference on Learning Representations}, 2023.

\bibitem[Liang et~al.(2024{\natexlab{a}})Liang, Cheng, Fan, Ling, Nie, Chen, Deng, Allen, Auerbach, Mahmood, et~al.]{liang2024quantifying}
Paul~Pu Liang, Yun Cheng, Xiang Fan, Chun~Kai Ling, Suzanne Nie, Richard Chen, Zihao Deng, Nicholas Allen, Randy Auerbach, Faisal Mahmood, et~al.
\newblock Quantifying \& modeling multimodal interactions: An information decomposition framework.
\newblock \emph{Advances in Neural Information Processing Systems}, 36, 2024{\natexlab{a}}.

\bibitem[Liang et~al.(2024{\natexlab{b}})Liang, Deng, Ma, Zou, Morency, and Salakhutdinov]{liang2024factorized}
Paul~Pu Liang, Zihao Deng, Martin~Q Ma, James~Y Zou, Louis-Philippe Morency, and Ruslan Salakhutdinov.
\newblock Factorized contrastive learning: Going beyond multi-view redundancy.
\newblock \emph{Advances in Neural Information Processing Systems}, 36, 2024{\natexlab{b}}.

\bibitem[Lin et~al.(2023)Lin, Li, Li, Li, and Gao]{lin2023diffusion}
Lequan Lin, Zhengkun Li, Ruikun Li, Xuliang Li, and Junbin Gao.
\newblock Diffusion models for time-series applications: a survey.
\newblock \emph{Frontiers of Information Technology \& Electronic Engineering}, pp.\  1--23, 2023.

\bibitem[Liu et~al.(2022)Liu, Wu, Wang, and Long]{liu2022non}
Yong Liu, Haixu Wu, Jianmin Wang, and Mingsheng Long.
\newblock Non-stationary transformers: Exploring the stationarity in time series forecasting.
\newblock \emph{Advances in Neural Information Processing Systems}, 35:\penalty0 9881--9893, 2022.

\bibitem[Liu et~al.(2023)Liu, Hu, Zhang, Wu, Wang, Ma, and Long]{liu2023itransformer}
Yong Liu, Tengge Hu, Haoran Zhang, Haixu Wu, Shiyu Wang, Lintao Ma, and Mingsheng Long.
\newblock itransformer: Inverted transformers are effective for time series forecasting.
\newblock In \emph{The Twelfth International Conference on Learning Representations}, 2023.

\bibitem[Lu et~al.(2023)Lu, Han, and Yang]{lu2023arm}
Jiecheng Lu, Xu~Han, and Shihao Yang.
\newblock Arm: Refining multivariate forecasting with adaptive temporal-contextual learning.
\newblock In \emph{The Twelfth International Conference on Learning Representations}, 2023.

\bibitem[Narasimhan et~al.(2024)Narasimhan, Agarwal, Akcin, Sanghavi, and Chinchali]{narasimhan2024time}
Sai~Shankar Narasimhan, Shubhankar Agarwal, Oguzhan Akcin, Sujay Sanghavi, and Sandeep Chinchali.
\newblock Time weaver: A conditional time series generation model.
\newblock \emph{arXiv preprint arXiv:2403.02682}, 2024.

\bibitem[Nie et~al.(2022)Nie, Nguyen, Sinthong, and Kalagnanam]{nie2022time}
Yuqi Nie, Nam~H Nguyen, Phanwadee Sinthong, and Jayant Kalagnanam.
\newblock A time series is worth 64 words: Long-term forecasting with transformers.
\newblock In \emph{The Eleventh International Conference on Learning Representations}, 2022.

\bibitem[Oord et~al.(2018)Oord, Li, and Vinyals]{oord2018representation}
Aaron van~den Oord, Yazhe Li, and Oriol Vinyals.
\newblock Representation learning with contrastive predictive coding.
\newblock \emph{arXiv preprint arXiv:1807.03748}, 2018.

\bibitem[Peebles \& Xie(2023)Peebles and Xie]{peebles2023scalable}
William Peebles and Saining Xie.
\newblock Scalable diffusion models with transformers.
\newblock In \emph{Proceedings of the IEEE/CVF International Conference on Computer Vision}, pp.\  4195--4205, 2023.

\bibitem[Rasul et~al.(2020)Rasul, Sheikh, Schuster, Bergmann, and Vollgraf]{rasul2020multivariate}
Kashif Rasul, Abdul-Saboor Sheikh, Ingmar Schuster, Urs~M Bergmann, and Roland Vollgraf.
\newblock Multivariate probabilistic time series forecasting via conditioned normalizing flows.
\newblock In \emph{International Conference on Learning Representations}, 2020.

\bibitem[Rasul et~al.(2021)Rasul, Seward, Schuster, and Vollgraf]{rasul2021autoregressive}
Kashif Rasul, Calvin Seward, Ingmar Schuster, and Roland Vollgraf.
\newblock Autoregressive denoising diffusion models for multivariate probabilistic time series forecasting.
\newblock In \emph{International Conference on Machine Learning}, pp.\  8857--8868. PMLR, 2021.

\bibitem[Salinas et~al.(2019)Salinas, Bohlke-Schneider, Callot, Medico, and Gasthaus]{salinas2019high}
David Salinas, Michael Bohlke-Schneider, Laurent Callot, Roberto Medico, and Jan Gasthaus.
\newblock High-dimensional multivariate forecasting with low-rank gaussian copula processes.
\newblock \emph{Advances in neural information processing systems}, 32, 2019.

\bibitem[Shen \& Kwok(2023)Shen and Kwok]{shen2023non}
Lifeng Shen and James Kwok.
\newblock Non-autoregressive conditional diffusion models for time series prediction.
\newblock In \emph{International Conference on Machine Learning}, pp.\  31016--31029. PMLR, 2023.

\bibitem[Shen et~al.(2023)Shen, Chen, and Kwok]{shen2023multi}
Lifeng Shen, Weiyu Chen, and James Kwok.
\newblock Multi-resolution diffusion models for time series forecasting.
\newblock In \emph{The Twelfth International Conference on Learning Representations}, 2023.

\bibitem[Song \& Ermon(2019)Song and Ermon]{song2019understanding}
Jiaming Song and Stefano Ermon.
\newblock Understanding the limitations of variational mutual information estimators.
\newblock In \emph{International Conference on Learning Representations}, 2019.

\bibitem[Song et~al.(2020)Song, Sohl-Dickstein, Kingma, Kumar, Ermon, and Poole]{song2020score}
Yang Song, Jascha Sohl-Dickstein, Diederik~P Kingma, Abhishek Kumar, Stefano Ermon, and Ben Poole.
\newblock Score-based generative modeling through stochastic differential equations.
\newblock \emph{arXiv preprint arXiv:2011.13456}, 2020.

\bibitem[Tashiro et~al.(2021)Tashiro, Song, Song, and Ermon]{tashiro2021csdi}
Yusuke Tashiro, Jiaming Song, Yang Song, and Stefano Ermon.
\newblock Csdi: Conditional score-based diffusion models for probabilistic time series imputation.
\newblock \emph{Advances in Neural Information Processing Systems}, 34:\penalty0 24804--24816, 2021.

\bibitem[Trirat et~al.(2024)Trirat, Shin, Kang, Nam, Na, Bae, Kim, Kim, and Lee]{trirat2024universal}
Patara Trirat, Yooju Shin, Junhyeok Kang, Youngeun Nam, Jihye Na, Minyoung Bae, Joeun Kim, Byunghyun Kim, and Jae-Gil Lee.
\newblock Universal time-series representation learning: A survey.
\newblock \emph{arXiv preprint arXiv:2401.03717}, 2024.

\bibitem[Tsai et~al.(2020)Tsai, Wu, Salakhutdinov, and Morency]{tsai2020self}
Yao-Hung~Hubert Tsai, Yue Wu, Ruslan Salakhutdinov, and Louis-Philippe Morency.
\newblock Self-supervised learning from a multi-view perspective.
\newblock In \emph{International Conference on Learning Representations}, 2020.

\bibitem[Wang \& Liu(2021)Wang and Liu]{wang2021understanding}
Feng Wang and Huaping Liu.
\newblock Understanding the behaviour of contrastive loss.
\newblock In \emph{Proceedings of the IEEE/CVF conference on computer vision and pattern recognition}, pp.\  2495--2504, 2021.

\bibitem[Wang et~al.(2024{\natexlab{a}})Wang, Han, Wang, and Zhang]{wang2024contrast}
Yihe Wang, Yu~Han, Haishuai Wang, and Xiang Zhang.
\newblock Contrast everything: A hierarchical contrastive framework for medical time-series.
\newblock \emph{Advances in Neural Information Processing Systems}, 36, 2024{\natexlab{a}}.

\bibitem[Wang et~al.(2023)Wang, Schiff, Gokaslan, Pan, Wang, De~Sa, and Kuleshov]{wang2023infodiffusion}
Yingheng Wang, Yair Schiff, Aaron Gokaslan, Weishen Pan, Fei Wang, Christopher De~Sa, and Volodymyr Kuleshov.
\newblock Infodiffusion: Representation learning using information maximizing diffusion models.
\newblock In \emph{International Conference on Machine Learning}, pp.\  36336--36354. PMLR, 2023.

\bibitem[Wang et~al.(2024{\natexlab{b}})Wang, Liu, Huang, Xiong, and Zhang]{wang2024multi}
Yongkang Wang, Xuan Liu, Feng Huang, Zhankun Xiong, and Wen Zhang.
\newblock A multi-modal contrastive diffusion model for therapeutic peptide generation.
\newblock In \emph{Proceedings of the AAAI Conference on Artificial Intelligence}, volume~38, pp.\  3--11, 2024{\natexlab{b}}.

\bibitem[Wang et~al.(2022)Wang, Xu, Zhang, Trajcevski, Zhong, and Zhou]{wang2022learning}
Zhiyuan Wang, Xovee Xu, Weifeng Zhang, Goce Trajcevski, Ting Zhong, and Fan Zhou.
\newblock Learning latent seasonal-trend representations for time series forecasting.
\newblock \emph{Advances in Neural Information Processing Systems}, 35:\penalty0 38775--38787, 2022.

\bibitem[Woo et~al.(2021)Woo, Liu, Sahoo, Kumar, and Hoi]{woo2021cost}
Gerald Woo, Chenghao Liu, Doyen Sahoo, Akshat Kumar, and Steven Hoi.
\newblock Cost: Contrastive learning of disentangled seasonal-trend representations for time series forecasting.
\newblock In \emph{International Conference on Learning Representations}, 2021.

\bibitem[Wu et~al.(2021)Wu, Xu, Wang, and Long]{wu2021autoformer}
Haixu Wu, Jiehui Xu, Jianmin Wang, and Mingsheng Long.
\newblock Autoformer: Decomposition transformers with auto-correlation for long-term series forecasting.
\newblock \emph{Advances in neural information processing systems}, 34:\penalty0 22419--22430, 2021.

\bibitem[Wu et~al.(2024)Wu, Luo, Kong, Papalexakis, and Steeg]{wu2024your}
Yunshu Wu, Yingtao Luo, Xianghao Kong, Evangelos~E Papalexakis, and Greg~Ver Steeg.
\newblock Your diffusion model is secretly a noise classifier and benefits from contrastive training.
\newblock \emph{arXiv preprint arXiv:2407.08946}, 2024.

\bibitem[Yang et~al.(2024)Yang, Jin, Wen, Zhang, Liang, Ma, Wang, Liu, Yang, Xu, et~al.]{yang2024survey}
Yiyuan Yang, Ming Jin, Haomin Wen, Chaoli Zhang, Yuxuan Liang, Lintao Ma, Yi~Wang, Chenghao Liu, Bin Yang, Zenglin Xu, et~al.
\newblock A survey on diffusion models for time series and spatio-temporal data.
\newblock \emph{arXiv preprint arXiv:2404.18886}, 2024.

\bibitem[Yoon et~al.(2019)Yoon, Jarrett, and Van~der Schaar]{yoon2019time}
Jinsung Yoon, Daniel Jarrett, and Mihaela Van~der Schaar.
\newblock Time-series generative adversarial networks.
\newblock \emph{Advances in neural information processing systems}, 32, 2019.

\bibitem[Yuan \& Qiao(2024)Yuan and Qiao]{yuan2024diffusion}
Xinyu Yuan and Yan Qiao.
\newblock Diffusion-ts: Interpretable diffusion for general time series generation.
\newblock \emph{arXiv preprint arXiv:2403.01742}, 2024.

\bibitem[Zeng et~al.(2023)Zeng, Chen, Zhang, and Xu]{zeng2023transformers}
Ailing Zeng, Muxi Chen, Lei Zhang, and Qiang Xu.
\newblock Are transformers effective for time series forecasting?
\newblock In \emph{Proceedings of the AAAI conference on artificial intelligence}, volume~37, pp.\  11121--11128, 2023.

\bibitem[Zhou et~al.(2021)Zhou, Zhang, Peng, Zhang, Li, Xiong, and Zhang]{zhou2021informer}
Haoyi Zhou, Shanghang Zhang, Jieqi Peng, Shuai Zhang, Jianxin Li, Hui Xiong, and Wancai Zhang.
\newblock Informer: Beyond efficient transformer for long sequence time-series forecasting.
\newblock In \emph{Proceedings of the AAAI conference on artificial intelligence}, volume~35, pp.\  11106--11115, 2021.

\bibitem[Zhou et~al.(2022)Zhou, Ma, Wen, Wang, Sun, and Jin]{zhou2022fedformer}
Tian Zhou, Ziqing Ma, Qingsong Wen, Xue Wang, Liang Sun, and Rong Jin.
\newblock Fedformer: Frequency enhanced decomposed transformer for long-term series forecasting.
\newblock In \emph{International conference on machine learning}, pp.\  27268--27286. PMLR, 2022.

\bibitem[Zhu et~al.(2022)Zhu, Wu, Olszewski, Ren, Tulyakov, and Yan]{zhu2022discrete}
Ye~Zhu, Yu~Wu, Kyle Olszewski, Jian Ren, Sergey Tulyakov, and Yan Yan.
\newblock Discrete contrastive diffusion for cross-modal music and image generation.
\newblock In \emph{The Eleventh International Conference on Learning Representations}, 2022.

\end{thebibliography}
